%% file: paper.tex
\documentclass[11pt]{article}
\usepackage{yub}

\usepackage[top=1in, bottom=1in, left=1in, right=1in]{geometry}
\usepackage{algorithmic}
\usepackage{algorithm}
\usepackage{subcaption}

\usepackage{hyperref}
\usepackage{cleveref}
\usepackage{url}

\hypersetup{
    colorlinks,
    linkcolor={blue!50!black},
    citecolor={blue!50!black},
    urlcolor=black,
}
\colorlet{linkequation}{blue}

\newcommand{\NNN}{N}
\newcommand{\cG}{\mathcal{G}}
\newcommand{\cF}{\mathcal{F}}
\newcommand{\poly}{\textup{poly}}

\newcommand\df{{W_{\mc{F}}}}
\newcommand\rw{{R_W}}
\newcommand\rb{{R_b}}
\newcommand\csigma{\kappa_\sigma}
\newcommand\bsigma{\beta_\sigma}
\newcommand\bw{B_w}
\newcommand{\pn}{\hat{p}^n}
\newcommand{\qn}{\hat{q}^n}

\newcommand{\ptail}{{p_{\rm tail}}}
\def\trainerror{\eps_{\rm train}}
\def\generror{\eps_{\rm gen}}
\def\dneuron{W_{\cF}}

\def\shownotes{0}  
\ifnum\shownotes=1
\newcommand{\authnote}[2]{$\ll$\textsf{\footnotesize #1 notes: #2}$\gg$}
\else
\newcommand{\authnote}[2]{}
\fi
\newcommand{\Tnote}[1]{{\color{red}\authnote{Tengyu}{#1}}}

\newcommand{\relu}{\textup{ReLU}}
\usepackage{tikz}
\usepackage{mathtools}

\newcommand*\circled[1]{\tikz[baseline=(char.base)]{
            \node[shape=circle,draw,inner sep=2pt] (char) {#1};}}

\newcommand{\Exp}{\mathop{\mathbb E}\displaylimits}

\title{Approximability of Discriminators Implies Diversity in GANs}

\author{Yu~Bai\thanks{Department of Statistics, Stanford
    University. \url{yub@stanford.edu}}
  \and
  Tengyu~Ma\thanks{Department of Computer Science and Department of
    Statistics, Stanford University. \url{tengyuma@stanford.edu}}
  \and
  Andrej~Risteski\thanks{MIT, Applied Mathematics and
    IDSS. \url{risteski@mit.edu}}}

\begin{document}
\maketitle

\newcommand{\fd}{cleantex-new}
\input{\fd/abstract.tex}
\input{\fd/intro.tex}

\input{\fd/prior-work.tex}
\input{\fd/prelim.tex}

\input{\fd/basic.tex}
\input{\fd/generator.tex}

\input{\fd/additive.tex}

\input{\fd/experiment-iclr.tex}
\input{\fd/conclusion.tex}

\subsection*{Acknowledgments}
The authors would like to thank Leon Bottou and John Duchi for
many insightful discussions.

\bibliographystyle{abbrvnat}
\bibliography{gan}

\appendix

\input{\fd/proof-prelim.tex}

\input{\fd/proof-basic.tex}

\input{\fd/proof-generator.tex}

\input{\fd/proof-additive.tex}

\input{\fd/appendix-experiment-iclr.tex}

\end{document}

%% file: cleantex-new/abstract.tex
\begin{abstract}
  While Generative Adversarial Networks (GANs) have empirically
  produced impressive results on learning complex real-world
  distributions, recent works have shown that they suffer from lack of
  diversity or mode collapse. The theoretical work of
  ~\citet{AroraGeLiMaZh17} suggests a dilemma about GANs' statistical
  properties: powerful discriminators cause overfitting, whereas weak
  discriminators cannot detect mode collapse.
  
  By contrast, we show in this paper that GANs can in principle learn
  distributions in Wasserstein distance (or KL-divergence in many
  cases) with polynomial sample complexity, if the discriminator class
  has strong distinguishing power against the particular generator
  class (instead of against all possible generators). For various
  generator classes such as mixture of Gaussians, exponential
  families, and invertible and injective neural networks generators,
  we design corresponding discriminators (which are often neural nets
  of specific architectures) such that the Integral Probability Metric
  (IPM) induced by the discriminators can provably approximate the
  Wasserstein distance and/or KL-divergence. This implies that if the
  training is successful, then the learned distribution is close to
  the true distribution in Wasserstein distance or KL divergence, and
  thus cannot drop modes. Our preliminary experiments show that on
  synthetic datasets the test IPM is well correlated with KL
  divergence or the Wasserstein distance, indicating that the lack of
  diversity in GANs may be caused by the sub-optimality in
  optimization instead of statistical inefficiency.
\end{abstract}
\Tnote{authonotes turned on!}

%% file: cleantex-new/intro.tex
\vspace{-0.25cm}
\section{Introduction}
In the past few years, we have witnessed great empirical success of Generative Adversarial Networks (GANs)~\citep{Goodfellow14} in generating high-quality samples in many domains. Various ideas have been proposed to further improve the quality of the learned distributions and the stability of the training. (See e.g., ~\citep{ArjovskyChBo17, odena2016conditional, xh2016SGAN, soumith2016DCGAN,tolstikhin2017adagan,Salimans2016ImprovedGANs,2016arXiv161204021J,2016arXiv161101673D, xu2017attngan} and the reference therein.)

However, understanding of GANs is still in its infancy. Do GANs
actually learn the target distribution?  Recent
work~\citep{AroraGeLiMaZh17,arora2018generalization,dumoulin2016adversarially}
has both theoretically and empirically brought the concern to light
that distributions learned by GANs suffer from mode collapse or lack
of diversity --- the learned distribution tends to miss a significant
amount of modes of the target distribution (elaborated in
Section~\ref{section:background}). The main message of this paper is
that the mode collapse can be in principle alleviated by designing
proper discriminators with strong distinguishing power against
specific families of generators such as special subclasses of neural
network generators (see Section~\ref{section:ra}
and~\ref{section:conjoined-design} for a detailed introduction.)

\subsection{Background on mode collapse in GANs}
\label{section:background}
We mostly focus on 
the Wasserstein GAN (WGAN)
formulation~\citep{ArjovskyChBo17} in this paper. 
Define the 
$\mc{F}$-Integral Probability Metric ($\cF$-IPM)~\citep{Muller97}
between distributions $p,q$ as
\begin{align}
  \label{equation:df-definition}
  W_{\mathcal{F}}(p, q) :=
  \sup_{f\in\mc{F}}\big|\E_{X\sim p}[f(X)] - \E_{X\sim q}[f(X)]\big|\,.
\end{align}
Given samples from distribution $p$, WGAN sets up a family of generators $\cG$, a
family of discriminators $\cF$, and aims to learn the data
distribution $p$ by solving 
\begin{align}
  \min_{q \in\mc{G}} ~~W_{\mc{F}}(\hat{p}^n, \hat{q}^m)\label{eqn:obj}\end{align}
 where $\hat{p}^n$ denotes ``the empirical version of the distribution $p$'', meaning the uniform distribution over a set of $n$ i.i.d samples from $p$ (and similarly $\hat{q}^m$.)
 
When $\mc{F}  = \{\textup{all 1-Lipschitz functions}\}$, IPM reduces
to the Wasserstein-1 distance $W_1$. In practice, parametric families
of functions $\cF$ such as multi-layer neural networks are used for
approximating Lipschitz functions, so that we can empirically optimize
this objective~\cref{eqn:obj} via gradient-based algorithms as long
as distributions in the family $\mc{G}$ have parameterized
samplers. (See Section~\ref{section:prelim} for more details.)
 
 One of the main theoretical and empirical concerns with GANs is the issue of
 ``mode-collapse''\citep{AroraGeLiMaZh17,Salimans2016ImprovedGANs} ---
 the learned distribution $q$
 tends to generate high-quality but low-diversity
 examples. Mathematically, the problem apparently arises from the fact that
IPM is weaker than $W_1$, and the mode-dropped
 distribution can fool the former~\citep{AroraGeLiMaZh17}: for a typical
 distribution $p$, there exists a distribution $q$ such that
 simultaneously the followings happen:
 \begin{align}\label{eqn:1}
W_{\mc{F}}(p,q) \lesssim \eps \textup{ and}~~W_{1}(p,q) \gtrsim 1.
 \end{align}
where $\lesssim, \gtrsim$ hide constant
 factors. In fact, setting $q = \hat{p}^N$ with $N = R(\mc{F})/\eps^2$, where $R(\mc{F})$ is a complexity measure of $\mc{F}$ (such as
 Rademacher complexity), $q$ satisfies~\cref{eqn:1} but is clearly a
 mode-dropped version of $p$ when $p$ has an exponential number of modes.

  Reasoning that the problem is with the strength of the discriminator,
 a natural solution is to increase it to larger families such as all
 1-Lipschitz functions. However, Arora et al.~\citep{AroraGeLiMaZh17}
 points out that Wasserstein-1 distance doesn't have good
 generalization properties: the empirical Wasserstein distance used in the
 optimization is very far from the population distance. Even for a
 spherical Gaussian distribution
 $p = \normal(0,\frac{1}{d}I_{d\times d})$ (or many other typical
 distributions), when the distribution $q$ is exactly equal to $p$, letting
 $\hat{q}^m$ and $\hat{p}^n$ be two empirical versions of $q$ and $p$
 with $m,n={\rm poly}(d)$, we have with high probability,
 \begin{align}
   W_{1}(\hat{p}^n, \hat{q}^m) \gtrsim 1~~~\textup{ even though
   }~~~W_{1}(p, q) =0.
 \end{align}
 Therefore even when learning succeeds ($p=q$), it cannot be gleaned from the
 empirical version of $W_1$.
 
 The observations above pose a dilemma in establishing the theories of
 GANs: powerful discriminators cause overfitting, whereas weak
 discriminators result in diversity issues because IPM doesn't
 approximate the Wasserstein distance
 The lack of diversity has also been
 observed empirically by~\citep{srivastava2017veegan, di2017max,
   borji2018pros, arora2018generalization}.

 \subsection{An approach to diversity: discriminator families
   with restricted approximability}
 \label{section:ra}
This paper proposes a resolution to the conundrum by designing a
discriminator class $\cF$ that is particularly strong against a
specific generator class $\cG$.
We say that a discriminator class
$\cF$ (and its IPM $W_{\cF}$) has {\it restricted approximability}
w.r.t. a generator class $\cG$ and the data distribution $p$, if $\cF$ can 
distinguish $p$ and any $q \in \cG$ approximately as
well as all 1-Lipschitz functions can do:
\begin{align}
  & \quad \textup{$W_{\cF}$ has restricted approximability
    w.r.t. $\cG$ and $p$} \nonumber \\
  & \triangleq ~\forall q \in
    \cG, ~\gamma_L(W_1(p, q))\lesssim W_{\cF}(p, q)\lesssim
    \gamma_U(W_1(p, q)), \label{eqn:conjoin}
\end{align}
where $\gamma_L(\cdot)$ and $\gamma_U(\cdot)$ are two monotone
nonnegative functions with $\gamma_L(0) = \gamma_U(0) = 0$. The paper
mostly focuses on $\gamma_{L}(t) = t^{\alpha}$ with $1\le\alpha\le 2$
and $\gamma_U(t) = t$, although we use the term ``restricted
approximability'' more
generally for this type of result (without tying it to a concrete
definition of $\gamma$). 
In other words, we are looking for
discriminators $\cF$ so that $\cF$-IPM can approximate the Wasserstein
distance $W_1$ for the data distribution $p$ and any $q\in\cG$.

Throughout the rest of this paper, we will focus on the realizable
case, that is, we assume $p\in\cG$, in which case we say
\emph{$W_{\cF}$ has restricted approximability with respect to $\cG$}
if~\cref{eqn:conjoin} holds for all $p,q\in\cG$. We note, however,
that such a framework allows the non-realizible case $p\notin\cG$ in
full generality (for example, results can be established through
designing $\cF$ that satisfies the requirement in
Lemma~\ref{thm:logp}).

A discriminator class $\cF$ with restricted approximability resolves
the dilemma in the following way. 

First, $\cF$ avoids mode collapse -- if the IPM between $p$ and $q$ is small, then by the left hand
side of~\cref{eqn:conjoin}, $p$ and $q$ are also close in Wasserstein distance and therefore 
significant
mode-dropping cannot happen. \footnote{Informally, if most of the modes of $p$ are $\eps$-far away from each other, then as long as $W_1(p,q)\ll \eps$, $q$ has to contain most of the modes of $p$.}

Second, we can pass from population-level guarantees to empirical-level guarantees -- 
as shown in~\citet{AroraGeLiMaZh17}, classical
capacity bounds such as the Rademacher complexity of $\cF$ relate
$\df(p,q)$ to $\df(\hat{p}^n, \hat{q}^m)$. Therefore, as long as the
capacity is bounded, we can expand on~\cref{eqn:conjoin} to get a full picture of the statistical
properties of Wasserstein GANs:
\begin{align}
  \forall q \in \cG, ~\gamma_L(W_1(p, q))\lesssim W_{\cF}(p,
  q)\approx W_{\mc{F}}(\hat{p}^n, \hat{q}^m)\lesssim \gamma_U(W_1(p,
  q)). \nonumber
\end{align}
Here the first inequality addresses the {\bf diversity 
  property} of the distance $W_{\cF}$, and the second approximation
addresses the {\bf generalization} of the distance, and the third
inequality provides the reverse guarantee that if the training fails to
find a solution with small IPM, then indeed $p$ and $q$ are far away in Wasserstein distance.\footnote{We
  also note that the third inequality can hold for all $p, q$ as long
  as $\cF$ is a subset of Lipschitz functions. } To the best of our
knowledge, this is the first theoretical framework  that tackles the statistical
theory of GANs with polynomial samples.

The main body of the paper will develop techniques for designing
discriminator class $\cF$ with restricted approximability for several
examples of generator classes including simple classes like mixtures of Gaussians,
exponential families, and more complicated classes like distributions generated by invertible neural
networks. In the next subsection, we will show that properly chosen
$\cF$ provides diversity guarantees such as
inequalities~\cref{eqn:conjoin}.

\subsection{Design of discriminators with restricted approximability}
\label{section:conjoined-design}

We start with relatively simple families of distributions $\cG$ such as
Gaussian distributions and exponential families, where we can directly
design $\cF$ to distinguish pairs of distribution in $\cG$. As we show
in Section~\ref{section:gaussian}, for Gaussians it
suffices to use one-layer neural networks with ReLU activations as
discriminators, and for exponential families to use linear combinations
of the sufficient statistics. 

In Section~\ref{section:generator}, we study the family of
distributions generated by invertible neural networks. We show that a
special type of neural network discriminators with one additional
layer than the generator has restricted approximability\footnote{This
  is consistent with the empirical finding that generators and
  discriminators with similar depths are often near-optimal choices of
  architectures.}. We show
this discriminator class guarantees that
$W_1(p,q)^{2}\lesssim \df(p,q) \lesssim W_1(p,q)$ where here we hide
polynomial dependencies on relevant parameters
(Theorem~\ref{theorem:invertible-generator}).  We remark that such
networks can also produce an exponentially large number of modes due to the
non-linearities, and our results imply that if $ \df(p,q)$ is small,
then most of these exponential modes will show up in the learned
distribution $q$. 


One limitation of the invertibility assumption is that it only
produces distributions supported on the entire space. The distribution of natural images is often believed to reside approximately on a low-dimensional manifold. When the distribution $p$ have a Lebesgue measure-zero support, the KL-divergence (or the reverse KL-divergence) is infinity unless the support of the estimated distribution coincides with the support of $p$.\footnote{The formal mathematical statement is that $\dkl(p\|q)$ is infinity unless $p$ is absolutely continuous with respect to $q$.} Therefore, while our proof makes crucial use of the KL-divergence in the invertible case, the KL-divergence is fundamentally not the proper measurement of the statistical distance for the cases where both $p$ and $q$ have low-dimensional supports. 

The crux of the technical part of the paper is to establish the approximation of Waserstein distance by IPMs for generators with low-dimensional supports. We will show that a variant of an IPM can still be sandwiched by
Wasserstein distance as in form of~\cref{eqn:conjoin} without
relating to KL-divergence
(Theorem~\ref{thm:additiveinforma}).  This demonstrates the
advantage of GANs over MLE approach on learning distributions with
low-dimensional supports. As the main proof technique, we develop tools for
approximating the log-density of a smoothed neural network generator.

We demonstrate in synthetic and controlled experiments that the IPM
correlates with the Wasserstein distance for low-dimensional
distributions with measure-zero support and correlates with KL-divergence
for the invertible generator family (where computation of KL is
feasible) (Section~\ref{section:exp-iclr}
and Appendix~\ref{sec:exp}.) The theory
suggests the possibility that when the KL-divergence or Wasserstein
distance is not measurable in more complicated settings, the test IPM
could serve as a candidate alternative for measuring the diversity and
quality of the learned distribution.  We also remark that on real
datasets, often the optimizer is tuned to carefully balance the
learning of generators and discriminators, and therefore the reported
training loss is often not the test IPM (which requires optimizing the
discriminator until optimality.) Anecdotally, the distributions
learned by GANs can often be distinguished by a well-trained
discriminator from the data distribution, which suggests that the IPM
is not well-optimized (See~\citet{lopez2016revisiting} for analysis
of for the original GANs formulation.) We conjecture that the lack of
diversity in real experiments may be caused by sub-optimality of the
optimization, rather than statistical inefficiency.

%% file: cleantex-new/prior-work.tex
\vspace{-0.2cm}
\subsection{Related work}
Various empirical proxy tests for diversity, memorization, and generalization
have been developed, such as interpolation between
images~\citep{soumith2016DCGAN}, semantic combination of images via
arithmetic in latent space~\citep{bojanowski2017optimizing},
classification tests~\citep{santurkar2017classification}, etc. These
results by and large indicate that while ``memorization'' is not an
issue with most GANs, lack of diversity frequently is.

As discussed thoroughly in the introduction,~\citet{AroraGeLiMaZh17,
  arora2018generalization} formalized the potential theoretical
sources of mode collapse from a weak discriminator, and proposed a
``birthday paradox'' that convincingly demonstrates this phenomenon is
real. Many architectures and algorithms have been proposed to remedy
or ameliorate mode collapse~\citep{dumoulin2016adversarially,
  srivastava2017veegan, di2017max, borji2018pros,LinKhFaOh17} with varying
success. \citet{feizi2017understanding} showed provable guarantees of
training GANs with quadratic discriminators when the generators are
Gaussians. However, to the best of our knowledge, there are no
provable solutions to this problem in more substantial generality.

The inspiring work of Zhang et al.~\citep{zhang2017discrimination} shows that the IPM is a proper metric (instead of a pseudo-metric) under a mild regularity condition. 
Moreover, it provides a KL-divergence bound with finite samples when
the densities of the true and estimated distributions exist. Our
Section~\ref{section:invertible-generator} can be seen as an extension
of~\citep[Proposition 2.9 and Corollary
3.5]{zhang2017discrimination}. The strength in our work is that we
develop statistical guarantees in Wasserstein distance for
distributions such as injective neural network generators, where the
data distribution resides on a low-dimensional manifold and thus
does not have proper density. 

~\citet{liang2017well} considers GANs in a non-parametric setup, one of the messages being that the sample complexity for learning GANs improves with the smoothness of the generator family.
However, the rate they derive is non-parametric -- exponential in the dimension -- unless the Fourier spectrum of the target family decays extremely fast, which can potentially be unrealistic in practical instances.

The invertible generator structure was used in
Flow-GAN~\citep{GroverDhEr18}, which observes that GAN training blows
up the KL on real dataset. Our theoretical result and
experiments show that successful GAN training (in terms of
the IPM) does imply learning in KL-divergence when the data
distribution can be generated by an invertible neural net. This suggests, along with the message in~\citep{GroverDhEr18}, that the real data cannot be generated by an invertible neural network. 
In addition, our theory implies that if the data can be generated by an \textit{injective} neural network (Section~\ref{s:additive}), we can bound the closeness between 
the learned distribution and the true distribution in Wasserstein distance (even though in this case, the KL divergence is no longer an informative measure for closeness.)


%% file: cleantex-new/prelim.tex
\section{Preliminaries and Notation}
\label{section:prelim}
The notion of IPM (recall the definition in~\cref{equation:df-definition}) includes
a number of statistical distances such as TV (total variation) and Wasserstein-1 distance by
taking $\mc{F}$ to be 1-bounded and 1-Lipschitz functions respectively. When $\mc{F}$
is a class of neural networks, we refer to the $\mc{F}$-IPM as
the {\it neural net IPM}.\footnote{This was defined as neural net distance in~\citep{AroraGeLiMaZh17}. }

There are many distances of interest between distributions that are not
IPMs, two of which we will particularly focus on: the KL
divergence $\dkl(p\|q)=\E_p[\log p(X)-\log q(X)]$ (when the densities
exist), and
the Wasserstein-2 distance, defined as
$W_2(p,q)^2 = \inf_{\pi\in\Pi} \E_{(X,Y)\sim\pi}[\norm{X-Y}^2]$ where
$\Pi$ be the set of couplings of $(p,q)$. We will only consider
distributions with finite second moments, so that $W_1$ and $W_2$
exist. 

\sloppy For any distribution $p$, we let $\hat{p}^n$ be the empirical
distribution of $n$ i.i.d. samples from $p$. The Rademacher complexity
of a function class $\cF$ on a distribution $p$ is
$R_n(\cF,p)=\E\left[\sup_{f\in\cF}|\frac{1}{n}\sum_{i=1}^n\eps_if(X_i)|\right]$
where $X_i \sim p$ i.i.d. and $\eps_i\sim \{\pm 1\}$ are
independent. We define
$R_n(\cF,\cG)=\sup_{p\in\cG}R_n(\cF,p)$ to be the largest Rademacher
complexity over $p\in\cG$. 
The training IPM loss (over the entire dataset) for the Wasserstein GAN,
assuming discriminator reaches optimality, is
$\E_{\hat{q}^n}\left[\df(\hat{p}^n, \hat{q}^n)\right]$\footnote{In the
  ideal case we can take the expectation over $q$, as the generator
  $q$ is able to generate infinitely many samples.}.
Generalization of the IPM is governed by the quantity
$R_n(\cF,\cG)$, as stated in the following result (see
Appendix~\ref{appendix:proof-generalization} for the proof): 
\begin{theorem}[Generalization, c.f.~\citep{AroraGeLiMaZh17}]
  \label{theorem:generalization}
  For any $p\in\mc{G}$, we have that 
  \begin{equation*}
    \forall q\in \cG, ~~\E_{\hat{p}^n}|\df(p, q) -
    \E_{\hat{q}^n}\left[\df(\hat{p}^n, \hat{q}^n)\right]| \leq
    4R_n(\cF,\cG).
  \end{equation*}
\end{theorem}

\noindent {\bf Miscellaneous notation.} We let $\normal(\mu,\Sigma)$ denote a
(multivariate) Gaussian distribution with mean $\mu$ and covariance
$\Sigma$. For quantities $a,b>0$ $a\lesssim b$ denotes that $a\le Cb$
for a universal constant $C>0$ unless otherwise stated explicitly.

%% file: cleantex-new/basic.tex
\section{Restricted Approximability for Basic Distributions}
\label{section:gaussian}

\subsection{Gaussian distributions}
\label{section:one-gaussian}
As a warm-up, we design discriminators with restricted approximability
for relatively simple parameterized distributions such Gaussian
distributions, exponential families, and mixtures of Gaussians. We
first prove that one-layer neural networks with ReLU
activation are strong enough to distinguish Gaussian distributions
with the restricted approximability guarantees.

We consider the set of Gaussian
distributions with bounded mean and well-conditioned covariance
$
  \cG = \set{p_\theta=\normal(\mu,\Sigma):\ltwo{\mu}\le
    D,\sigma_{\min}^2I_d \preceq
    \Sigma\preceq \sigma_{\max}^2I_d } \label{eqn:gaussian}
$.
Here $D, \sigma_{\min}$ and $\sigma_{\max}$ are considered as given
hyper-parameters. We will show that the IPM $W_{\cF}$ induced by the
following discriminators has restricted approximability w.r.t. $\cG$:  
\begin{equation}
  \mc{F} \defeq \set{x\mapsto \relu(v^\top
    x+b):\ltwo{v}\le 1, |b|\le D}, \label{eqn:gaussianf}
\end{equation}
\begin{theorem}
  \label{theorem:one-gaussian}
  The set of one-layer neural networks ($\cF$ defined in
  ~\cref{eqn:gaussianf}) has restricted approximability w.r.t. the
  Gaussian distributions in $\cG$ in the sense
  that for any $p,q\in \cG$
  \begin{equation*}
    \kappa\cdot W_1(p, q) \lesssim \dneuron(p, q) \le W_1(p, q).
  \end{equation*}
  with
  $\kappa = \frac{1}{\sqrt{d}} \frac{\sigma_{\min}}{\sigma_{\max}}$.
  Moreover, $\cF$ satisfies Rademacher complexity bound
  $ R_n(\mc{F},\cG)\lesssim \frac{D+\sigma_{\max}\sqrt{d}}{\sqrt{n}}$.
\end{theorem}
Apart from absolute constants, the lower and upper bounds differ by a
factor of $1/\sqrt{d}$.\footnote{As shown
in~\citep{feizi2017understanding}, the optimal discriminator for
Gaussian distributions are quadratic functions. } We point out that
the $1/\sqrt{d}$ factor is not improvable unless using functions more
sophisticated than Lipschitz functions of one-dimensional projections
of $x$. Indeed, $\df(p,q)$ is upper bounded by the maximum Wasserstein
distance between one-dimensional projections of $p,q$, which is on the
order of $W_1(p,q)/\sqrt{d}$ when $p,q$ have spherical covariances.
The proof is deferred to Section~\ref{appendix:proof-one-gaussian}.

\noindent {\bf Extension to mixture of Gaussians.} Discriminator
family $\mc{F}$ with restricted approximability can also be designed
for mixture of Gaussians. We defer this result and the proof to
Appendix~\ref{appendix:mixture-of-gaussians}.

\subsection{Exponential families}
\label{section:exponential-family}
Now we consider exponential families and show that the linear
combinations of the sufficient statistics are a family of
discriminators with restricted approximability. Concretely, let
$\mc{G}=\{p_{\theta}:\theta\in \Theta\subset\R^k\}$ be an exponential
family, where
$
  p_\theta(x) = \frac{1}{Z(\theta)}\exp(\<\theta, T(x)\> ), ~\forall
  x\in \mc{X}\subset \R^d \label{eqn:exp_family} 
$: here $T:\R^d \to\R^k$ is the vector of sufficient statistics, and
$Z(\theta)$ is the partition function. Let the discriminator family be
all linear functionals over the features $T(x)$:
$
  \mc{F} = \set{x \to \<v,T(x)\>: \ltwo{v}\le 1}.
$
\begin{theorem}
  \label{theorem:exponential-family}
  Let $\cG$ be the exponential family and $\cF$ be the discriminators defined above. Assume that the log partition function
  $\log Z(\theta)$ satisfies that $\gamma I\preceq \nabla^2 \log
  Z(\theta)\preceq \beta I$. Then we have for any $p,q\in \cG$, 
  \begin{equation}
    \frac{\gamma}{\sqrt{\beta}}\sqrt{\dkl(p\|q)}
    \le \df(p,q) \le
    \frac{\beta}{\sqrt{\gamma}}\sqrt{\dkl(p\|q)}. \label{eqn:exp_kl}
  \end{equation}
  If we further assume $\mc{X}$ has diameter $D$ and $T(x)$ is
  $L$-Lipschitz in $\mc{X}$. Then, 
  \begin{align}
    \frac{D\gamma}{\sqrt{\beta}}W_1(p,q) \lesssim \df(p,q)
    \le L \cdot W_1(p,q) \label{eqn:exp_w}
  \end{align}
  Moreover, $\cF$ has Rademacher complexity bound
  $ R_n(\cF, \cG)\le
  \sqrt{\frac{\sup_{\theta\in\Theta}E_{p_{\theta}}[\ltwo{T(X)}^2]}{n}}$.
\end{theorem}
We note that the log partition function $\log Z(\theta)$ is always
convex, and therefore our assumptions only require in addition that
the curvature (i.e. the Fisher information matrix) has a strictly
positive lower bound and a global upper bound. For the
bound~\cref{eqn:exp_w}, some geometric assumptions on the sufficient
statistics are necessary because the Wasserstein distance
intrinsically depends on the underlying geometry of $x$, which are not
specified in exponential families by default. The proof
of~\cref{eqn:exp_kl} follows straightforwardly from the standard
theory of exponential families. The proof of~\cref{eqn:exp_w} requires
machinery that we will develop in Section~\ref{section:generator} and
is therefore deferred to
Section~\ref{appendix:proof-exponential-family}.

%% file: cleantex-new/generator.tex
\section{Restricted Approximability for Neural Net Generators}
\label{section:generator}
In this section, we design discriminators with restricted
approximability for neural net generators, a family of distributions
that are widely used in GANs to model real data.


In Section~\ref{section:invertible-generator} we consider the
invertible neural networks generators which have proper densities. In
Section~\ref{s:additive}, we extend the results to the more general
and challenging setting of injective neural networks generators, where
the latent variables are allowed to have lower dimension than the
observable dimensions (Theorem~\ref{thm:additiveinforma}) and the
distributions no longer have densities.

\subsection{Invertible neural network generators}
\label{section:invertible-generator}
In this section, we consider the generators that are parameterized by
invertible neural networks\footnote{Our techniques also
  applies to other parameterized invertible generators but for
  simplicity we only focus on neural networks.}.
Concretely, let $\mathfrak{G}$ be a family of neural networks
$\mathfrak{G} = \set{G_\theta:\theta\in\Theta}$. Let $p_\theta$ be the
distribution of
\begin{equation}
  \label{model:invertible-generator}
  X=G_\theta(Z),~~Z\sim\normal(0, \diag(\gamma^2)).
\end{equation}
where $G_\theta$ is a neural network with parameters $\theta$ and
$\gamma\in \R^d$ standard deviation of hidden factors. By allowing the
variances to be non-spherical, we allow each hidden dimension to have
a different impact on the output distribution. In particular, the case
$\gamma=[\ones_{k}, \delta\ones_{d-k}]$ for some $\delta\ll 1$ has the
ability to model data around a ``$k$-dimensional manifold'' with some
noise on the level of $\delta$.

We are interested in the set of invertible neural networks
$G_\theta$. 
We let our family $\mc{G}$ consist
of standard $\ell$-layer feedforward nets $x=G_\theta(z)$ of the form
\begin{equation*}
  x = W_\ell\sigma(W_{\ell-1}\sigma(\cdots\sigma(W_1z + b_1)\cdots) +
  b_{\ell-1}) + b_\ell,
\end{equation*}
where $W_i\in\R^{d\times d}$ are invertible, $b_i\in\R^d$, and
$\sigma:\R\to\R$ is the activation function, on which we make the
following assumption:
\begin{assumption}[Invertible generators]
  \label{assumption:invertible-generator}
  Let $R_W, R_b, \csigma, \beta_\sigma>0, \delta\in(0,1]$ be
  parameters which are considered as constants (that may depend on the
  dimension). We consider neural networks $G_{\theta}$ that are
  parameterized by parameters $\theta=(W_i,b_i)_{i\in[\ell]}$ belonging
  to the set
  \begin{equation*}
    \Theta = \set{(W_i, b_i)_{i\in[\ell]}:~\max\set{\opnorm{W_i},
        \opnorm{W_i^{-1}}} \le \rw,~\ltwo{b_i}\le \rb,~\forall
      i\in[\ell]}.
  \end{equation*}
  The activation function $\sigma$ is twice-differentiable
  with $\sigma(0)=0$,
  $\sigma'(t)\in[\csigma^{-1},1]$, and
  $|(\sigma^{-1})''/(\sigma^{-1})'| \le \bsigma$. The standard
  deviation of the hidden factors satisfy $\gamma_i\in[\delta, 1]$.
\end{assumption}
Clearly, such a neural net is invertible, and
its inverse is also a feedforward neural net with activation
$\sigma^{-1}$. We note that a smoothed version of Leaky
ReLU~\citep{xu2015empirical}
satisfies all the conditions on the activation functions.
Further, it is necessary to impose some assumptions on the generator
networks because arbitrary neural networks are likely to be able to
implement pseudo-random functions which can't be distinguished from
random functions by even any polynomial time algorithms.

\begin{lemma}
  \label{lemma:logp-neural-network}
  For any $\theta\in\Theta$, the function $\log p_\theta$ can be
  computed by a neural network with at most $\ell+1$ layers,
  $O(\ell d^2)$ parameters, and activation function among
  $\{\sigma^{-1},\log\sigma^{-1'},(\cdot)^2\}$ of the form
  \begin{equation}
    \label{eqn:F-1}
    f_\phi(x) = \frac{1}{2}\<h_1, \diag(\gamma^{-2})h_1\> +
    \sum_{k=2}^{\ell}\<\ones_d, \log\sigma^{-1'}(h_j)\> + C,
  \end{equation}
  where $h_\ell=W_\ell(x-b_\ell)$,
  $h_k=W_k(\sigma^{-1}(h_{k+1})-b_k)$ for $k\in\{\ell-1,\dots,1\}$,
  and the parameter $\phi=((W_j,b_j)_{j=1}^\ell, C)$ satisfies
  $\phi\in\Phi=\{\phi:\opnorm{W_j}\le\rw,~\ltwo{b_j}\le \rb,~|C|\le
  (\ell-1)d\log\rw\}$.
  As a direct
  consequence, the following family $\cF$ of neural networks
  with activation functions above of at most $\ell+2$ layers
  contains all the functions
  $
  \set{\log p - \log q:~p,q\in \cG}:
  $
  \begin{align}
    \label{eqn:F-2}
    \mathcal{F} = \set{f_{\phi_1}-f_{\phi_2}:\phi_1,\phi_2\in\Phi}.
  \end{align}
    \end{lemma}

We note that the exact form of the parameterized family $\cF$ is likely not very important in practice, since other family of neural nets also possibly contain good approximations of $\log p -\log q$ (which can be seen partly from experiments in Section~\ref{sec:exp}.)

The proof builds on the change-of-variable formula
$\log
p_\theta(x)=\log\phi_\gamma(G_\theta^{-1}(x))+\log|\det\frac{\partial
  G_\theta^{-1}(x)}{\partial x}|$ (where $\phi_\gamma$ is the density
of $Z\sim\normal(0, \diag(\gamma^2))$) and the observation that
$G_\theta^{-1}$ is a feedforward neural net with $\ell$ layers. Note
that the log-det of the Jacobian involves computing the determinant of
the (inverse) weight matrices. A priori such computation is
non-trivial for a given $G_\theta$.
However, it's just some constant that does not depend on the input,
therefore it can be 
representable by adding a bias on the final output layer. This frees us from further structural assumptions on the weight matrices (in contrast to the architectures in flow-GANs~\citep{GulrajaniAhArDuCo17}).
We defer the proof of Lemma~\ref{lemma:logp-neural-network} to
Section~\ref{appendix:proof-logp-neural-network}.

\begin{theorem}
  \label{theorem:invertible-generator}
  Suppose $\mc{G}=\set{p_\theta:\theta\in\Theta}$ is the set of
  invertible-generator distributions as defined
  in~\cref{model:invertible-generator} satisfying
  Assumption~\ref{assumption:invertible-generator}. Then, the
  discriminator class $\cF$ defined in
  Lemma~\ref{lemma:logp-neural-network} has restricted approximability
  w.r.t. $\cG$ in the sense that for any $p,q\in\mc{G}$,
  \begin{align*}
    & W_1(p,q)^2 \lesssim \dkl(p\|q)+ \dkl(q\|p) \leq\df(p,q) \lesssim
      \frac{\sqrt{d}}{\delta^2}\left(W_1(p,q) + d\exp(-10d)\right),
              \end{align*}
  When $n\gtrsim \max\set{d, \delta^{-8}\log 1/\delta}$, we have the
  generalization bound
  $ R_n(\mc{F}, \mc{G}) \le \generror \defeq
  \sqrt{\frac{d^4\log n}{\delta^4 n}}$.
\end{theorem}

The proof of Theorem~\ref{theorem:invertible-generator} uses the
following lemma that relates the KL divergence to the IPM when the log
densities exist and belong to the family of discriminators.
\begin{lemma}[{Special case of \citep[Proposition 2.9]{zhang2017discrimination}}]
  \label{thm:logp}
  Let $\eps > 0$. Suppose $\cF$ satisfies that for every $q\in \cG$,
  there exists $f\in \cF$ such that $\|f -(\log p -\log q)\|_\infty\le
  \epsilon$, and that all the functions in $\cF$ are
  $L$-Lipschitz. Then,  
  \begin{align}
    \dkl(p\|q) + \dkl(q\|p) -\eps\le \df(p,q) \le L\cdot W_1(p,q).
  \end{align}
\end{lemma}
We outline a proof sketch of
Theorem~\ref{theorem:invertible-generator} below and defer the full
proof to Appendix~\ref{section:proof-invertible-generator}. As we
choose the discriminator class as in
Lemma~\ref{lemma:logp-neural-network} which implements
$\log p-\log q$ for any $p,q\in\cG$, by Lemma~\ref{thm:logp},
$\df(p,q)$ is lower bounded by $\dkl(p\|q)+\dkl(q\|p)$. It thus
suffices to (1) lower bound this quantity by the Wasserstein distance
and (2) upper bound $\df(p,q)$ by the Wasserstein distance.

To establish (1), we will prove in
Lemma~\ref{lemma:invertible-generator-lower-bound} that
for any $p,q\in\mc{G}$,
\begin{equation*}
  W_1(p,q)^2 \le W_2(p,q)^2 \lesssim \dkl(p\|q) + \dkl(q\|p).
\end{equation*}
Such a result is the simple implication of {\it transportation
  inequalities} by Bobkov-G{\"o}tze and Gozlan
(Theorem~\ref{theorem:kl-ge-wasserstein}), which state that if $X\sim p$ (or $q$) and
$f$ is $1$-Lipschitz implies that $f(X)$ is sub-Gaussian, then the inequality above holds. In our invertible generator
case, we have $X=G_\theta(Z)$ where $Z$ are independent Gaussians, so
as long as $G_\theta$ is suitably Lipschitz, $f(X)=f(G_\theta(Z))$ is
a sub-Gaussian random variable by the standard Gaussian concentration
result~\citep{Vershynin10}.

The upper bound (2) would have been immediate if functions in $\mc{F}$ are
Lipschitz globally in the whole space. While this is not strictly true, we give two workarounds --
by either doing a truncation argument to get a $W_1$ bound with some
tail probability, or a $W_2$ bound which only requires the Lipschitz
constant to grow at most linearly in $\ltwo{x}$. This is done in
Theorem~\ref{theorem:df-le-wasserstein} as a straightforward extension
of the result in~\citep{PolyanskiyWu16}.

Combining the restricted approximability and the generalization bound,
we immediately obtain that if the training succeeds with small
expected IPM (over the randomness of the learned distributions), then
the estimated distribution $q$ is close to the true distribution $p$
in Wasserstein distance.  
\begin{corollary}
	In the setting of Theorem~\ref{theorem:invertible-generator}, 
with high probability over the choice of
training data $\hat{p}^n$, we have that if the training process returns a
distribution $q\in \cG$ such that 
$
\E_{\hat{q}^n}[\df(\hat{p}^n, \hat{q}^n)] \le \trainerror
$, 
then with  $\generror \defeq
\sqrt{\frac{d^4\log n}{\delta^4 n}}$, we have
\begin{align}
W_1(p, q) \le W_2(p,q) \lesssim (\trainerror + \generror)^{1/2}.
\end{align}

\end{corollary}
We note that the training error is measured by $\E_{\hat{q}^m}[\df(\hat{p}^n, \hat{q}^m)]$, the expected IPM over the randomness of the learned distributions, which is a measurable value because one can draw fresh samples from $q$ to estimate the expectation.
It's an important open question to design efficient algorithms to achieve a small training error according to this definition, and this is left for future work.

%% file: cleantex-new/additive.tex
\subsection{Injective neural network generators}
\label{s:additive}
In this section we consider injective neural network generators
(defined below) which generate distributions residing on a low
dimensional manifold. This is a more realistic setting than
Section~\ref{section:invertible-generator} for modeling real images,
but technically more challenging because the KL divergence becomes
infinity, rendering Lemma~\ref{thm:logp} useless. Nevertheless, we
design a novel divergence between two distributions that is sandwiched
by Wasserstein distance and can be optimized as IPM.

Concretely, we consider a family of neural net generators
$\mathfrak{G} = \set{G_\theta:\R^k \rightarrow \R^d}$ where  $k < d$ and
$G_{\theta}$ is injective function. \footnote{In other words,
  $G_{\theta}(x)\neq G_{\theta}(y)$ if $x\neq y$. } Therefore,
$G_{\theta}$ is invertible only on the image of $G_{\theta}$, which is
a $k$-dimensional manifold in $\R ^d$.  Let $\cG$ be the corresponding
family of distributions generated by neural nets in $\mathfrak{G}$.

Our key idea is to design a variant of the IPM, which provably approximates the Wasserstein distance. 
Let
$p^{\beta}$ denote the convolution of the distribution $p$ with a Gaussian distribution
$\normal(0,\beta^2I)$.
We define a smoothed $\mc{F}$-IPM between $p,q$ as
\begin{align}
  \tilde{d}_{\cF}(p,q) \triangleq \inf_{\beta\ge 0} ~ (\df(p^\beta,
  q^\beta) + \beta\log 1/\beta)^{1/2},
\end{align}

Clearly $\tilde{d}_{\cF}$ can be optimized as $W_{\cF}$ with an additional variable $\beta$ introduced in the optimization. We show that for certain discriminator class (see Section~\ref{sec:proof:additive} for the details of the construction) such that $\tilde{d}_{\cF}$ approximates the Wasserstein distance. 
\begin{theorem}[Informal version of
  Theorem~\ref{theorem:additive-noise}]
  \label{thm:additiveinforma}
  Let $\cG$ be defined as above. There exists a discriminator class $\cF$ such that 
  for any pair of distributions $p,q\in \cG$, we have
  \begin{align}
    W_1(p,q)	\lesssim \tilde{d}_{\cF}(p, q) \lesssim \poly(d)\cdot W_1(p,q)^{1/6} + \exp(-\Omega(d)).
  \end{align}
	Furthermore, when $n \gtrsim \mbox{poly}(d)$, 
	we have the generalization bound 
  \begin{equation*}
    R_n(\mc{F}, \cG) \lesssim \poly(d) \sqrt{\frac{\log
        n}{n}}
  \end{equation*}
  Here $\poly(d)$ hides polynomial dependencies on $d$ and several other parameters that will be defined in the formal version (Theorem~\ref{theorem:additive-noise}.)
\end{theorem}
The direct implication of the theorem is that if $\tilde{d}(\hat{p}^n, \hat{q}^n)$ is small for $n \gtrsim \poly(d)$, then $W(p,q)$ is  guaranteed to be also small and thus we don't have mode collapse.

%% file: cleantex-new/experiment-iclr.tex
\section{Simulation}
\label{section:exp-iclr}

Our theoretical results on neural network generators in
Section~\ref{section:generator} convey the message that mode collapse
will not happen as long as the discriminator family $\cF$ has
restricted approximability with respect to the generator family
$\cG$. In particular, the IPM $W_{\cF}(p,q)$ is upper and lower
bounded by the Wasserstein distance $W_1(p,q)$ given the restricted
approximability. We design certain specific discriminator classes in
our theory to guarantee this, but we suspect it holds more generally
in GAN training in practice.

We perform two sets of synthetic experiments to
confirm that the practice is indeed consistent with our theory.
We design synthetic datasets, set up
suitable generators, and train GANs with either our theoretically
proposed discriminator class with restricted approximability, or
vanilla neural network discriminators of reasonable capacity. In both
cases, we show that IPM is well correlated with the Wasserstein / KL
divergence, suggesting that the restricted approximability may indeed
hold in practice. This suggests that the difficulty of GAN training in
practice may come from the optimization difficulty rather than
statistical inefficiency, as we observe evidence of good statistical
behaviors on ``typcial'' discriminator classes.


We briefly describe the experiments here and defer details of the second experiment to Appendix~\ref{sec:exp}.
\begin{enumerate}[(a)]
\item We learn synthetic 2D datasets with neural net generators and
  discriminators and show that the IPM is well-correlated with the
  Wasserstein distance (Section~\ref{section:toy-experiment}).
\item We learn invertible neural net generators with discriminators of
  restricted approximability and vanilla architectures
  (Appendix~\ref{sec:exp}). We show that the IPM is well-correlated
  with the KL divergence, both along training and when we consider two
  generators that are perturbations of each other (the purpose of the
  latter being to eliminate any effects of the optimization).
\end{enumerate}

\subsection{Experiments on Synthetic 2d Datasets}
\label{section:toy-experiment}

In this section, we perform synthetic experiments with WGANs that
learn various curves in two dimensions. In particular, we will train
GANs that learn the unit circle and a ``swiss roll''
curve~\citep{GulrajaniAhArDuCo17} -- both distributions are supported
on a one-dimensional manifold in $\R^2$, therefore the KL divergence
does not exist, but one can use the Wasserstein distance to measure
the quality of the learned generator.

We show that WGANs are able
to learn both distributions pretty well, and the IPM $W_{\cF}$ is
strongly correlated with the Wasserstein distance $W_1$. These ground
truth distributions are not covered in our
Theorems~\ref{theorem:invertible-generator}
and~\ref{thm:additiveinforma}, but our results show evidence that
restricted approximability is still quite likely to hold here.

{\bf Ground truth distributions}
We set the ground truth distribution to be a unit circle or a Swiss
roll curve, sampled from
\begin{align*}
  & \textrm{Circle}:~(x, y)\sim{\rm Uniform}(\set{(x, y):x^2+y^2=1}) \\
  & \textrm{Swiss roll}:~(x, y) = (z\cos(4\pi z), z\sin(4\pi z)):~z\sim{\rm Uniform}([0.25, 1]).
\end{align*}

{\bf Generators and discriminators} We use standard
two-hidden-layer ReLU nets as both the generator class and the
discriminator class. The generator architecture is 2-50-50-2, and the
discriminator architecture is 2-50-50-1. We use the RMSProp
optimizer~\citep{TielemanHi12} as our update rule, the learning rates
are $10^{-4}$ for both the generator and discriminator, and we perform
10 steps on the discriminator in between each generator step.

{\bf Metric} We compare two metrics between the ground truth
distribution $p$ and the learned distribution $q$ along training:
\begin{enumerate}[(1)]
\item The neural net IPM $W_{\cF}(p, q)$, computed on fresh batches
  from $p,q$ through optimizing a separate discriminator from cold
  start.
\item The Wasserstein distance $W_1(p,q)$, computed on fresh batches
  from $p,q$ using the {\tt POT}
  package\footnote{\url{https://pot.readthedocs.io/en/stable/index.html}}.
  As data are in two dimensions, the empirical Wasserstein distance
  $W_1(\hat{p}, \hat{q})$ does not suffer from the curse of
  dimensionality and is a good proxy of the true Wasserstein distance
  $W_1(p,q)$~\citep{WeedBa17}.
\end{enumerate}



\begin{figure}[h!]
  \centering
  \begin{subfigure}[b]{0.29\textwidth}
    \includegraphics[width=\textwidth]{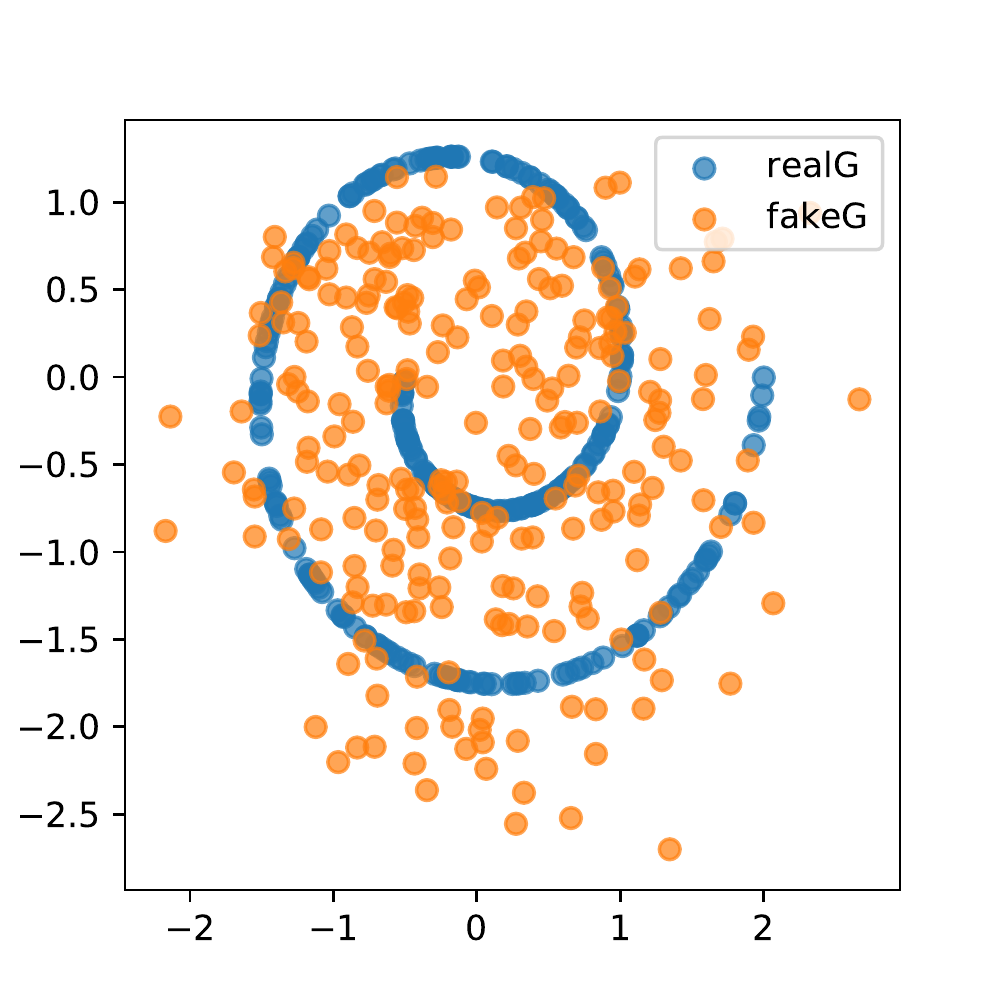}
    \caption{\small Iteration 500.}
    \label{figure:swiss1}
  \end{subfigure}
  ~
  \begin{subfigure}[b]{0.29\textwidth}
    \includegraphics[width=\textwidth]{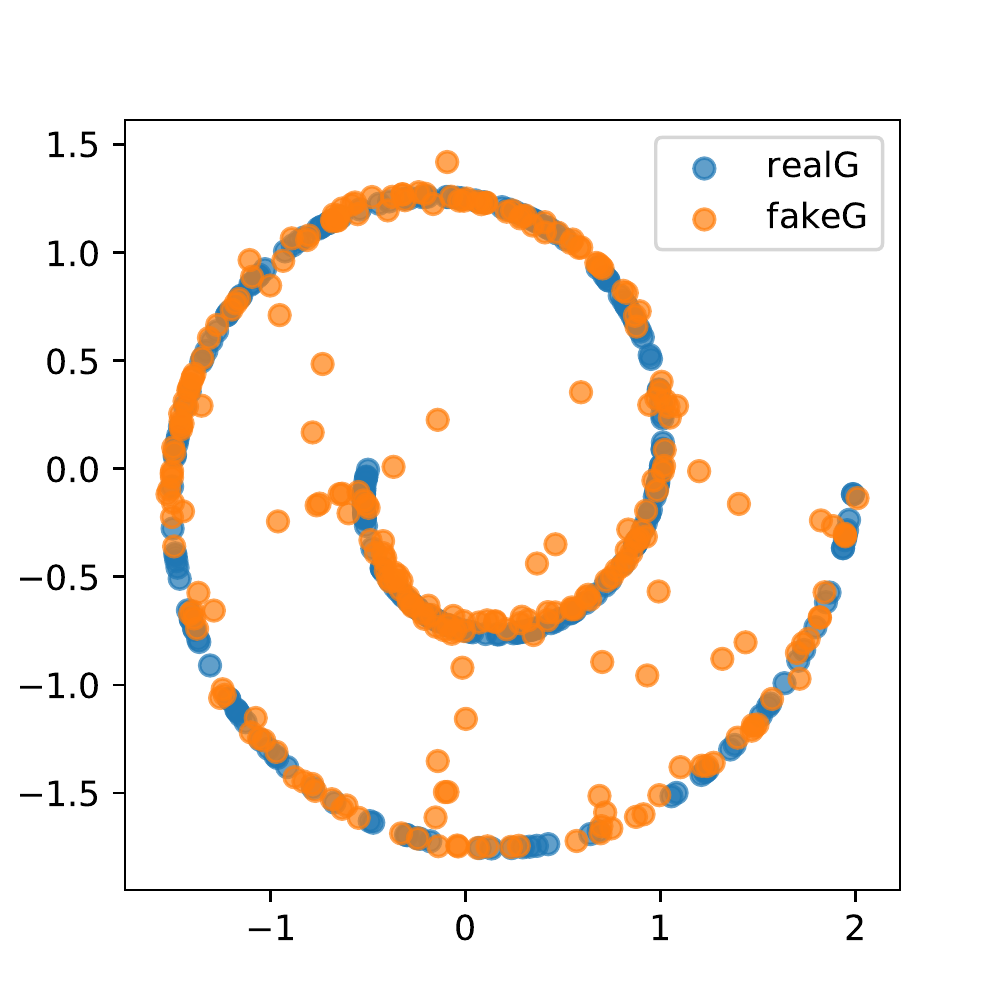}
    \caption{\small Iteration 10000.}
    \label{figure:swiss2}
  \end{subfigure}
  ~
  \begin{subfigure}[b]{0.35\textwidth}
    \includegraphics[width=\textwidth]{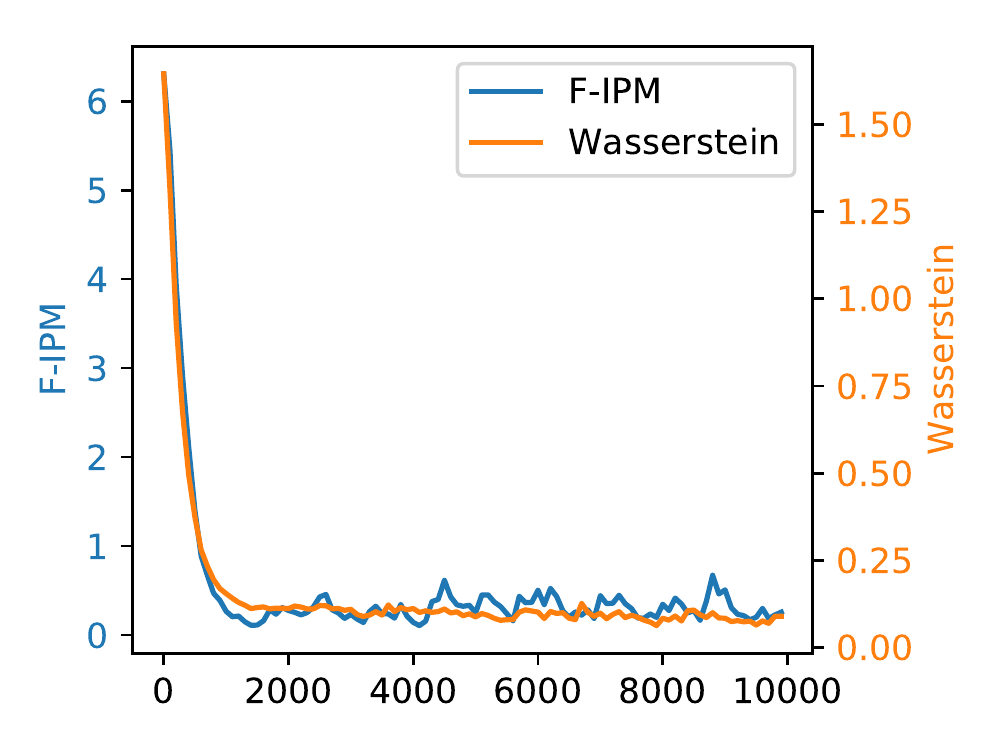}
    \caption{\small Comparing IPM and Wasserstein.}
    \label{figure:swiss3}
  \end{subfigure}
  \caption{\small Experiments on the swiss roll dataset. The neural
    net IPM, the Wasserstein distance, and the sample quality are
    correlated along training. (a)(b): Sample batches from the ground
    truth and the learned generator at iteration 500 and 5000. (c):
    Comparing the F-IPM and the Wasserstein distance. RealG and fakeG
    denote the ground truth generator and the learned generator,
    respectively.}
  \label{figure:swiss}
\end{figure}

{\bf Result}
See Figure~\ref{figure:swiss} for the Swiss roll experiment and
Figure~\ref{figure:circle} (in Appendix~\ref{appendix:toy-experiment})
for the unit circle experiment. On both datasets, the learned
generator is very close to the ground truth distribution at iteration
10000. Furthermore, the neural net IPM and the Wasserstein distance
are well correlated. At iteration 500, the generators have not quite
learned the true distributions yet (by looking at the sampled
batches), and the IPM and Wasserstein distance are indeed large.

%% file: cleantex-new/conclusion.tex
\section{Conclusion}
We present the first polynomial-in-dimension sample complexity bounds
for learning various distributions (such as Gaussians, exponential
families, invertible neural networks generators) using GANs with
convergence guarantees in Wasserstein distance (for distributions with
low-dimensional supports) or KL divergence. The analysis technique
proceeds via designing discriminators with restricted approximability
-- a class of discriminators tailored to the generator class in
consideration which have good generalization and mode collapse
avoidance properties.  

We hope our techniques can be in future extended to other families of
distributions with tighter sample complexity bounds. This would entail
designing discriminators that have better restricted approximability
bounds, and generally exploring and generalizing approximation theory
results in the context of GANs. We hope such explorations will prove
as rich and satisfying as they have been in the vanilla functional
approximation settings. 

%% file: cleantex-new/proof-prelim.tex
\section{Proofs for Section~\ref{section:prelim}}
\subsection{Proof of Theorem~\ref{theorem:generalization}}
\label{appendix:proof-generalization}
Fixing $\pn$, consider a random sample $\qn$. It is easy to
verify that the $\mc{F}$-IPM satisfies the triangle inequality,
so we have
\begin{align*}
  & \quad |\df(p,q) - \E_{\qn}[\df(\pn,\qn)]| \le
    \E_{\qn}[|\df(p,q) - \df(\pn,\qn)] \\
  & \le \E_{\qn}[\df(p,\pn) + \df(q,\qn)] =
      \df(p,\pn) + \E_{\qn}[\df(q,\qn)].
\end{align*}
Taking expectation over $\pn$ on the above bound yields
\begin{equation*}
  \E_{\pn}\left[|\df(p,q) - \E_{\qn}[\df(\pn,\qn)]|\right] \le
  \E_{\pn}[\df(p,\pn)] + \E_{\qn}[\df(q,\qn)].
\end{equation*}
So it suffices to bound $\E_{\pn}[\df(p,\pn)]$ by $2R_n(\cF,\cG)$ and
the same bound will hold for $q$. Let $X_i$ be the samples in
$\pn$. By symmetrization, we have 
\begin{equation*}
  \df(p,\pn) = \E\left[\sup_{f\in\mc{F}} \left|\frac{1}{n} \sum_{i=1}^n f(X_i) -
      \E_p[f(X)]\right|\right]
  \le 2\E\left[\sup_{f\in\mc{F}} \left| \frac{1}{n}\sum_{i=1}^n
      \eps_if(X_i)\right|\right] = 2\E[R_n(\mc{F}, p)] \le
  2R_n(\cF,\cG).
\end{equation*}
Adding up this bound and the same bound for $q$ gives the desired
result.

%% file: cleantex-new/proof-basic.tex
\section{Proofs for Section~\ref{section:gaussian}}
\subsection{Proof of Theorem~\ref{theorem:one-gaussian}}
\label{appendix:proof-one-gaussian}
Recall that our discriminator family is
\begin{equation*}
  \mc{F} = \set{x\mapsto \sigma(v^\top x+b):\ltwo{v}\le 1, |b|\le D}.
\end{equation*}

\paragraph{Restricted approximability}
The upper bound $\df(p_1,p_2)\le W_1(p_1,p_2)$ follows directly from
the fact that functions in $\mc{F}$ are 1-Lipschitz.

We now establish the lower bound.  First, we recover the mean
distance, in which we use the following
simple fact: a linear discriminator is the sum of two ReLU
discriminators, or mathematically $t=\sigma(t)-\sigma(-t)$. Taking
$v=\frac{\mu_1-\mu_2}{\ltwo{\mu_1-\mu_2}}$, we have
\begin{eqnarray*}
  && \ltwo{\mu_1 - \mu_2} = v^\top \mu_1 - v^\top \mu_2 =
     \E_{p_1}[v^\top X] - \E_{p_2}[v^\top X] \\
  &=& \left( \E_{p_1}[\sigma(v^\top X)] - \E_{p_2}[\sigma(v^\top X)]
      \right)  + \left( -\E_{p_1}[\sigma(-v^\top X)] +
      \E_{p_2}[\sigma(-v^\top X)] \right) \\
  &\le& \left| \E_{p_1}[\sigma(v^\top X)] - \E_{p_2}[\sigma(v^\top X)]
        \right| + \left| \E_{p_1}[\sigma(-v^\top X)] -
        \E_{p_2}[\sigma(-v^\top X)] \right|.
\end{eqnarray*}
Therefore at least one of the above two terms is greater than
$\ltwo{\mu_1-\mu_2}/2$, which shows that
$\dneuron(p_1,p_2)\ge\ltwo{\mu_1-\mu_2}/2$.

For the covariance distance, we need to actually compute
$\E_p[\sigma(v^\top X+b)]$ for $p=\normal(\mu,\Sigma)$.
Note that $X\eqnd \Sigma^{1/2}Z+\mu$, where
$Z\sim\normal(0,I_d)$. Further, we have
$v^\top X\eqnd \ltwo{\Sigma^{1/2} v}W + v^\top\mu$ for
$W\sim\normal(0,1)$, therefore
\begin{align*}
  & \quad \E_p[\sigma(v^\top X + b)] =
  \E\left[\sigma\left(\ltwo{\Sigma^{1/2}v}W+v^\top\mu+b\right)\right]
  \\
  & = \ltwo{\Sigma^{1/2}v} 
    \E\left[
    \sigma\left(W+\frac{v^\top\mu+b}{\ltwo{\Sigma^{1/2}v}}\right)
    \right] = \ltwo{\Sigma^{1/2}v}
    R\left(\frac{v^\top\mu+b}{\ltwo{\Sigma^{1/2}v}}\right).
\end{align*}
(Defining $R(a)=\E[\max\set{W+a, 0}]$ for $W\sim\normal(0,1)$.)
Therefore, the neuron distance between the two Gaussians is
\begin{equation*}
  \dneuron(p_1, p_2) = \sup_{\ltwo{v}\le 1, |b|\le D}
  \left|\ltwo{\Sigma_1^{1/2}v} 
    R\left(\frac{v^\top\mu_1+b}{\ltwo{\Sigma_1^{1/2}v}}\right) -
    \ltwo{\Sigma_2^{1/2}v} 
    R\left(\frac{v^\top\mu_2+b}{\ltwo{\Sigma_2^{1/2}v}}\right)\right|,
\end{equation*}
As $a\mapsto\max\set{a+w, 0}$ is strictly increasing for all $w$, the
function $R$ is strictly increasing. It is also a basic fact that
$R(0)=1/\sqrt{2\pi}$.

Consider any fixed $v$. By
flipping the sign of $v$, we can let $v^\top\mu_1\ge v^\top\mu_2$
without changing $\ltwo{\Sigma_i^{1/2}v}$. Now, letting
$b=-v^\top(\mu_1-\mu_2)/2$ (note that $|b|\le D$ is a valid choice),
we have
\begin{equation*}
  v^\top\mu_1 + b = \frac{v^\top(\mu_1-\mu_2)}{2} \ge
  0,~~v^\top\mu_2 + b = -\frac{v^\top(\mu_1-\mu_2)}{2} \le 0.
\end{equation*}
As $R$ is strictly increasing, for this choice of $(v,b)$ we have
\begin{align*}
  & \quad \ltwo{\Sigma_1^{1/2}v} 
    R\left(\frac{v^\top\mu_1+b}{\ltwo{\Sigma_1^{1/2}v}}\right) -
    \ltwo{\Sigma_2^{1/2}v} 
    R\left(\frac{v^\top\mu_2+b}{\ltwo{\Sigma_2^{1/2}v}}\right) \\
  & \ge
    R(0)\left(\ltwo{\Sigma_1^{1/2}v} - \ltwo{\Sigma_2^{1/2}v}\right)
    = \frac{1}{\sqrt{2\pi}} \left(\ltwo{\Sigma_1^{1/2}v} -
    \ltwo{\Sigma_2^{1/2}v}\right).
\end{align*}
Ranging over $\ltwo{v}\le 1$ we then have
\begin{equation*}
  \dneuron(p_1, p_2) \ge \frac{1}{\sqrt{2\pi}} \sup_{\ltwo{v}\le 1}
  \left| \ltwo{\Sigma_1^{1/2}v} - \ltwo{\Sigma_2^{1/2}v} \right|.
\end{equation*}
The quantity in the supremum can be further bounded as 
\begin{equation*}
  \left| \ltwo{\Sigma_1^{1/2}v} - \ltwo{\Sigma_2^{1/2}v} \right| =
  \frac{|v^\top(\Sigma_1-\Sigma_2)v|}{\ltwo{\Sigma_1^{1/2}v} +
    \ltwo{\Sigma_2^{1/2}v}} \ge
  \frac{|v^\top(\Sigma_1-\Sigma_2)v|}{\lambdamax{\Sigma_1^{1/2}}
    +\lambdamax{\Sigma_2^{1/2}}}.
\end{equation*}
Choosing $v=\vmax{\Sigma_1-\Sigma_2}$ gives
\begin{equation*}
  \dneuron(p_1, p_2) \ge \frac{1}{\sqrt{2\pi}}\sup_{\ltwo{v}\le 1}
  \left| \ltwo{\Sigma_1^{1/2}v} - \ltwo{\Sigma_2^{1/2}v} \right| \ge
  \frac{\opnorm{\Sigma_1-\Sigma_2}}{\sqrt{2\pi}2\sigma_{\max}}.
\end{equation*}
Now, using the perturbation bound
\begin{equation*}
  \opnorm{\Sigma_1^{1/2} - \Sigma_2^{1/2}} \le
  \frac{1}{\lambda_{\min}(\Sigma_1) + \lambda_{\min}(\Sigma_2)} \cdot
  \opnorm{\Sigma_1 - \Sigma_2} \le
  \frac{1}{2\sigma_{\min}}\opnorm{\Sigma_1 - \Sigma_2},
\end{equation*}
(cf.~\citep[Lemma 2.2]{Schmitt92}), we get
\begin{equation*}
  \dneuron(p_1, p_2) \ge \frac{1}{2\sqrt{2\pi}\sigma_{\max}} \cdot
  2\sigma_{\min}\opnorm{\Sigma_1^{1/2} - \Sigma_2^{1/2}} \ge
  \frac{\sigma_{\min}}{\sqrt{2\pi}\sigma_{\max}\sqrt{d}}
  \lfro{\Sigma_1^{1/2} - \Sigma_2^{1/2}}.
\end{equation*}
Combining the above bound with the bound in the mean difference, we
get
\begin{eqnarray}
  && \dneuron(p_1, p_2) \ge \frac{1}{2}\left(
     \frac{\ltwo{\mu_1-\mu_2}}{2} +
     \frac{\sigma_{\min}}{\sqrt{2\pi d}\sigma_{\max}}\lfro{\Sigma_1^{1/2}
     - \Sigma_2^{1/2}} \right) \nonumber \\
  &\ge& \frac{\sigma_{\min}}{2\sqrt{2\pi d}\sigma_{\max}}
        \sqrt{\ltwo{\mu_1-\mu_2}^2 + \inf_{U^\top
        U=UU^\top=I_d}\lfro{\Sigma_1^{1/2} - U\Sigma_2^{1/2}}^2}
        \nonumber \\
  &=& \frac{\sigma_{\min}}{2\sqrt{2\pi d}\sigma_{\max}} \cdot W_2(p_1,
      p_2) \ge \frac{\sigma_{\min}}{2\sqrt{2\pi d}\sigma_{\max}}\cdot
      W_1(p_1, p_2) \label{equation:gaussian-df-ge-w2}
\end{eqnarray}
The last equality following directly from the closed-form expression
of the $W_2$ distance between two Gaussians~\citep[Proposition
3]{MasarottoPaZe18}. Thus the claimed lower bound holds with
$c=1/(2\sqrt{2\pi})$.

\paragraph{KL Bound}
We use the $W_2$ distance to bridge the KL and the
$\mc{F}$-distance, which uses the machinery developed in
Section~\ref{appendix:proof-generator}. Let $p_1,p_2$ be two Gaussians
distributions with parameters $\theta_i=(\mu_i,\Sigma_i)\in\Theta$. By
the equality
\begin{equation*}
  \dkl(p_1\|p_2) + \dkl(p_2\|p_1) = (\E_{p_1}[\log p_1(X)] -
  \E_{p_2}[\log p_1(X)]) + (\E_{p_2}[\log p_2(X)] - \E_{p_1}[\log p_2(X)]),
\end{equation*}
it suffices to upper bound the term only involving $\log p_1(X)$ (the
other follows similarly), which by
Theorem~\ref{theorem:df-le-wasserstein} requires bounding the growth
of $\ltwo{\grad \log p_1(x)}$. We have
\begin{equation*}
  \ltwo{\grad \log p_1(x)} = \ltwo{\Sigma_1^{-1}(x-\mu_1)} \le
  \sigma_{\min}^{-2}\ltwo{x-\mu_1}.
\end{equation*}
Further $\E_{p_i}[\ltwo{x-\mu_1}^2] \le
\tr(\Sigma_i)+\ltwo{\mu_i-\mu_1}^2\le d\sigma_{\max}^2+4D^2$ for
$i=1,2$, therefore by (a trivial variant of)
Theorem~\ref{theorem:df-le-wasserstein}(c) we get
\begin{equation*}
  \E_{p_1}[\log p_1(X)] - \E_{p_2}[\log p_1(X)] \le
  \sigma_{\min}^{-2}(\sqrt{d}\sigma_{\max}+2D)W_2(p_1,p_2).
\end{equation*}
The same bound holds for $\log p_2$. Adding them up and substituting
the bound~\cref{equation:gaussian-df-ge-w2} gives that
\begin{equation*}
  \dkl(p_1\|p_2) + \dkl(p_2\|p_1) \lesssim \frac{\sqrt{d}\sigma_{\max}
  + 2D}{\sigma_{\min}^2} W_2(p_1, p_2) \lesssim
\frac{\sqrt{d}\sigma_{\max}(\sqrt{d}\sigma_{\max} +
  D)}{\sigma_{\min}^3} \df(p_1,p_2).
\end{equation*}

\paragraph{Generalization}
We wish to bound for all $\theta=(\mu,\Sigma)\in\Theta$
\begin{equation*}
  R_n(\mc{F}, p_\theta) = \E_{p_\theta}\left[ \sup_{\ltwo{v}\le
      1,|b|\le D} \left| \frac{1}{n}\sum_{i=1}^n\eps_i\sigma(v^\top
      X_i+b) \right|\right].
\end{equation*}
As $\sigma:\R\to\R$ is 1-Lipschitz, by the Rademacher contraction
inequality~\citep{LedouxTa13}, we have
\begin{equation*}
  \E_{p_\theta}\left[ \sup_{\ltwo{v}\le 1,|b|\le D} \left|
      \frac{1}{n}\sum_{i=1}^n\eps_i\sigma(v^\top X_i+b) \right|\right]
  \le 2 \E_{p_\theta}\left[ \sup_{\ltwo{v}\le
      1,|b|\le D} \left| \frac{1}{n}\sum_{i=1}^n\eps_i(v^\top
      X_i+b) \right|\right].
\end{equation*}
The right hand side can be bounded directly as
\begin{eqnarray*}
  && \E_{p_\theta}\left[ \sup_{\ltwo{v}\le
     1,|b|\le D} \left| \frac{1}{n}\sum_{i=1}^n\eps_i(v^\top
     X_i+b) \right|\right] = E_{p_\theta}\left[ \sup_{\ltwo{v}\le
     1,|b|\le D}
     \left|(b+v^\top\mu)\frac{1}{n}\sum_{i=1}^n\eps_i +
     \frac{1}{n}\sum_{i=1}^n \eps_iv^\top(X_i-\mu)\right|\right] \\
  &\le& \sup_{\ltwo{v}\le 1,|b|\le
        D}|b+v^\top\mu|
        \E\left[\left|\frac{1}{n}\sum_{i=1}^n\eps_i\right|\right] +
        \E_{p_\theta}\left[\sup_{\ltwo{v}\le 1}\left|\<v,
        \frac{1}{n}\sum_{i=1}^n\eps_i(X_i-\mu)\>\right|\right] \\
  &\le& 2D \E\left[\left|\frac{1}{n}\sum_{i=1}^n\eps_i\right|\right] +
        \E_{p_\theta}\left[\ltwo{ 
        \frac{1}{n}\sum_{i=1}^n\eps_i(X_i-\mu)} \right] \\
  &\le&
        2D\sqrt{\E\left[\left(\frac{1}{n}\sum_{i=1}^n\eps_i\right)^2\right]}
        + \sqrt{\E_{p_\theta}\left[\ltwo{ 
        \frac{1}{n}\sum_{i=1}^n\eps_i(X_i-\mu)}^2 \right]} \\
  &=& \frac{2D+\sqrt{\tr(\Sigma)}}{\sqrt{n}} \le
      \frac{2D+\sigma_{\max}\sqrt{d}}{\sqrt{n}}. 
\end{eqnarray*}

\subsection{Proof of Theorem~\ref{theorem:exponential-family}}
\label{appendix:proof-exponential-family}
\paragraph{KL bounds}
Recall the basic property of exponential family that 
$ A(\theta) = \E_{p_\theta}[T(X)] $.
Suppose $p=p_{\theta_1}$ and $q=p_{\theta_2}$. 
Then, 
\begin{align}
\df(p, q)
& = \sup_{\ltwo{v}  \le
	1}\E_{p_{\theta_1}}[\<v, T(X)\>] - \E_{p_{\theta_2}}[\<v,
T(X)\>] \nonumber\\
& = \sup_{\ltwo{v}\le 1} \<v, \grad A(\theta_1)-\grad A(\theta_2)\>
= \ltwo{\grad A(\theta_1) - \grad A(\theta_2)}.\nonumber
\end{align}
By the assumption on $\nabla^2 A$ we have that 
\begin{align}
\gamma \|\theta_1-\theta_2\|_2\le \df(p_{\theta_1}, p_{\theta_2})\le
\beta \|\theta_1-\theta_2\|_2 \label{eqn:4}
\end{align}
Moreover, the exponential family also satisfies that 
\begin{equation*}
\dkl(p_{\theta_1}\|p_{\theta_2}) = A(\theta_2) - A(\theta_1) -
\<\grad A(\theta_1), \theta_2-\theta_1\> = \int_{0}^1 \rho^\top
\nabla^2 A(\theta_2 + t\rho)\rho dt 
\end{equation*}
where $\rho = \theta_1-\theta_2$. Using the assumption we have that
$\gamma \|\theta_1-\theta_2\|^2 \le \rho^\top \nabla^2 A(\theta_2 +
t\rho)\rho \le \beta \|\theta_1-\theta_2\|^2$ and therefore $\frac 1 2
\gamma \|\theta_1-\theta_2\|^2 \le\dkl(p_{\theta_1}\|p_{\theta_2})
\le \frac 1 2\beta \|\theta_1-\theta_2\|^2$.  Combining this with
~\cref{eqn:4} we complete the proof.

\paragraph{Wasserstein bounds}
We show~\cref{eqn:exp_w}. As ${\rm diam}(\mc{X})=D$, there
exists $x_0\in\mc{X}$ such that $\norm{x-x_0}\le D$ for all
$x\in\mc{X}$. Hence for any 1-Lipschitz function $f:\R^d\to\R$ we have
that $|f(X)-f(x_0)|\le\ltwo{x-x_0}\le D$. By the Hoeffding Lemma,
$f(X)$ is $D^2/4$-sub-Gaussian. Applying
Theorem~\ref{theorem:kl-ge-wasserstein}(a), we get that for any $p,q\in\cG$,
\begin{equation*}
  W_1(p,q) \le \sqrt{\frac{D^2}{2}\dkl(p\|q)} \lesssim
  \frac{D\gamma}{\sqrt{\beta}}\cdot \df(p,q).
\end{equation*}

\paragraph{Generalization} For any $\theta\in\Theta$ we compute the
Rademacher complexity
\begin{eqnarray*}
	&& R_n(\mc{F}, p_\theta) = \E_{p_\theta}\left[\sup_{\ltwo{v}\le 1}
	\left| \frac{1}{n}\sum_{i=1}^n\eps_i\<v, T(X_i)\> \right| \right] =
	\E_{p_\theta}\left[ \ltwo{\frac{1}{n}\sum_{i=1}^n\eps_iT(X_i)}
	\right] \\
	&\le& \sqrt{\E_{p_\theta}\left[
		\ltwo{\frac{1}{n}\sum_{i=1}^n\eps_iT(X_i)}^2 \right]} =
	\sqrt{\frac{E_{p_\theta}[\ltwo{T(X)}^2]}{n}}.
\end{eqnarray*}

\section{Results on Mixture of Gaussians}
\label{appendix:mixture-of-gaussians}
We consider mixture of $k$ identity-covariance Gaussians on $\R^d$:
\begin{equation*}
 p_{\theta} = \sum_{i=1}^k w_i\normal(\mu_i, I_d),~~~w_i\ge
 0,~\sum_{i=1}^k w_i = 1.
\end{equation*}
We assume that $\theta\in\Theta=\set{\ltwo{\mu_i}\le D,w_i\ge
 \exp(-\bw)~i\in[k]}$.

We will use a one-hidden-layer neural network that implements (a
slight modification of) $\log p_\theta$:
\begin{equation*}
 \mc{F}= \set{ f_1 - f_2: f_i =
   \log\sum_{j=1}^kw_j^{(i)}\exp\left(\mu_j^{(i)\top} x + 
     b_j^{(i)}\right) : \exp(-\bw) \le w_j^{(i)} \le
   1,~\ltwo{\mu_j^{(i)}}\le D,~0\ge b_j^{(i)}\ge -D^2}.
\end{equation*}
\begin{theorem}
 \label{theorem:mixture-of-gaussians}
 The family $\mc{F}$ is suitable for learning mixture of $k$
 Gaussians. Namely, we have that
 \begin{enumerate}[(1)]
 \item (Restricted approximability) For any
   $\theta_1,\theta_2\in\Theta$, we have
   \begin{equation*}
     \frac{1}{D^2+1} \cdot W_1^2(p_{\theta_1}, p_{\theta_2}) \le
     \df(p_{\theta_1}, p_{\theta_2}) \le 2D\cdot W_1(p_{\theta_1},
     p_{\theta_2}).
   \end{equation*}
 \item (Generalization) We have for some absolute constant $C>0$ that
   \begin{equation*}
     \sup_{\theta\in\Theta}R_n(\mc{F},
     p_{\theta})\le C\sqrt{\frac{k(\log k+D^2+\bw)d\log n}{n}}.
   \end{equation*}
 \end{enumerate}
\end{theorem}

\subsection{The Gaussian concentration result}
The Gaussian concentration result~\citep[Proposition
5.34]{Vershynin10} will be used here and in later proofs, which we
provide for convenience.
\begin{lemma}[Gaussian concentration]
  \label{lemma:gaussian-concentration}
  Suppose $X\sim\normal(0,I_d)$ and $f:\R^d\to\R$ is $L$-Lipschitz,
  then $f(X)$ is $L^2$-sub-Gaussian. 
\end{lemma}

\subsection{Proof of Theorem~\ref{theorem:mixture-of-gaussians}}
\paragraph{Restricted approximability}
For the upper bound, it suffices to show that each
\begin{equation}
  \label{equation:mog-net}
  f(x) = \log\sum_{j=1}^k w_j\exp\left(\mu_j^\top x + b_j\right)
\end{equation}
is $D$-Lipschitz. Indeed, we have
\begin{equation*}
  \ltwo{\grad\log f(x)} = \ltwo{\frac{\sum_{j=1}^kw_j\exp(\mu_j^\top
      x+b_j)\mu_j}{\sum_{j=1}^kw_j\exp(\mu_j^\top x + b_j)}} \le
  \frac{\sum_{j=1}^kw_j\exp(\mu_j^\top 
    x+b_j)\ltwo{\mu_j}}{\sum_{j=1}^kw_j\exp(\mu_j^\top x + b_j)} \le
  D.
\end{equation*}
This further shows that every discriminator $f_1-f_2\in\mc{F}$ is at
most $2D$-Lipschitz, so by Theorem~\ref{theorem:df-le-wasserstein}(a)
we get the upper bound.

We now establish the lower bound. As $\mc{F}$ implements the KL
divergence, for any two $p_1,p_2\in\mc{P}$, we have
\begin{equation*}
  \df(p_1, p_2) \ge \dkl(p_1\|p_2) + \dkl(p_2\|p_1).
\end{equation*}
We consider regularity properties of the distributions $p_1,p_2$ in
the Bobkov-Gotze sense
(Theorem~\ref{theorem:kl-ge-wasserstein}(a)). Suppose $p_1=\sum
w_j\normal(\mu_j,I_d)$. For any 1-Lipschitz function $f:\R^d\to\R$, we
have
\begin{equation*}
  f(X) \eqnd \sum_{j=1}^j f(\normal(\mu_j,I_d)).
\end{equation*}
Letting $X_j\sim\normal(\mu_j,I_d)$ be the mixture components. By the
Gaussian concentration (Lemma~\ref{lemma:gaussian-concentration}),
each $f(X_j)$ is $1$-sub-Gaussian, so we have for any $\lambda\in\R$
\begin{eqnarray*}
  \E[e^{\lambda f(X)}] = \sum_{j=1}^k w_j \E[e^{\lambda f(X_j)}] \le
  \sum_{j=1}^k e^{\lambda\E[f(X_j)]+\lambda^2/2} =
  e^{\lambda^2/2}\underbrace{\sum_{j=1}^k
  w_je^{\lambda\E[f(X_j)]}}_{\rm I}. 
\end{eqnarray*}
Now, term I is precisely the MGF of a discrete random
variable on $Y\in\R$ which takes value $\E[f(X_j)]$ with probability
$w_j$. For $Z\sim\normal(0,1)$ we have
\begin{equation*}
  |\E[f(X_j)] - \E[f(Z)]| = |\E[f(\mu_j+Z)] - \E[f(Z)]| \le
  \E[|f(\mu_j+Z)-f(Z)|] \le \ltwo{\mu_j}\le D.
\end{equation*}
Therefore the values $\set{\E[f(X_j)]}_{j\in[k]}$ lie in an interval
of length at most $2D$. By the Hoeffding's Lemma, $Y$ is
$D^2$-sub-Gaussian, so we have ${\rm I}\le
\exp(\lambda\E[Y]+D^2\lambda^2/2)$, and so
\begin{equation*}
  \E[e^{\lambda f(X)}] \le \exp\left(\frac{\lambda^2}{2} +
    \lambda\E[Y] + D^2\lambda^2/2\right) = \exp\left(\lambda\E[Y] +
    \frac{\lambda^2(D^2+1)}{2}\right).
\end{equation*}
Therefore $f(X)$ is at most $(D^2+1)$-sub-Gaussian, and thus $X$
satisfies the Bobkov-Gozlan condition with $\sigma^2=D^2+1$. Applying
Theorem~\ref{theorem:kl-ge-wasserstein}(a) we get
\begin{equation*}
  \df(p_1,p_2) \ge \dkl(p_1\|p_2) + \dkl(p_2\|p_1) \ge
  \frac{1}{D^2+1}\cdot W(p_1,p_2).
\end{equation*}

\paragraph{Generalization}
Reparametrize the one-hidden-layer neural net~\cref{equation:mog-net}
as
\begin{equation*}
  f_\theta(x) = \log\sum_{j=1}^k\exp(\mu_j^\top x + \underbrace{b_j +
    \log w_j}_{c_j}).
\end{equation*}
It then suffices to bound the Rademacher complexity of $f_\theta$ for
$\theta\in\Theta=\set{\ltwo{\mu_j}\le D,c_j\in[-(D^2+\bw),0]}$. Define
the metric
\begin{equation*}
  \rho(\theta, \theta') = \max_{j\in[k]}
  \max\set{\ltwo{\mu_j-\mu_j'},|c_j-c_j'|}
\end{equation*}
and the Rademacher process
\begin{equation*}
  Y_\theta = \frac{1}{n}\sum_{i=1}^n \eps_if_\theta(X_i) =
  \frac{1}{n}\sum_{i=1}^n \eps_i\log\sum_{j=1}^k \exp(\mu_j^\top X_i +
  c_j),
\end{equation*}
we show that $Y_\theta$ is suitably Lipschitz in $\theta$ (in the
$\rho$ metric) and use a one-step discretization bound. Indeed, we
have
\begin{equation*}
  \ltwo{\grad_{\mu_j}Y_\theta} = \ltwo{\frac{1}{n}\sum_{i=1}^n
    \eps_i\frac{\exp(\mu_j^\top X_i +
      c_j)}{\sum_{j=1}^k\exp(\mu_j^\top X_i + c_j)}X_i} \le
  \frac{1}{n}\sum_{i=1}^n \ltwo{X_i}
\end{equation*}
and
\begin{equation*}
  |\grad_{c_j}Y_\theta| = \left|
    \frac{1}{n}\sum_{i=1}^n\eps_i\frac{\exp(\mu_j^\top X_i + 
      c_j)}{\sum_{j=1}^k\exp(\mu_j^\top X_i + c_j)} \right| \le 1.
\end{equation*}
Therefore, for any $\eps>0$ we have
\begin{equation}
  \label{equation:mog-discretization-error}
  \E\left[ \sup_{\rho(\theta,\theta')\le\eps} |Y_\theta -
    Y_{\theta'}| \right] \le Ck\left( \E\left[\ltwo{X_1}\right] +1
  \right)\eps \le Ck(D+\sqrt{d})\eps
\end{equation}
for some constant $C>0$.

We now bound the expected supremum of the max over a covering set. Let
$\mc{N}(\Theta,\rho,\eps)$ be a $\eps$-covering set of $\Theta$ under
$\rho$, and $N(\Theta,\rho,\eps)$ be the covering number. As $\rho$
looks at each $\mu_i,c_j$ separately, its covering number can be upper
bounded by the product of each separate covering:
\begin{equation*}
  N(\Theta,\rho,\eps) \le \prod_{j=1}^k N(\ball_2(D),
  \ltwo{\cdot},\eps)
  \cdot N([-(D^2+\bw),0],|\cdot|,\eps) \le \exp\left(
    kd\log\frac{3D}{\eps} + k\log\frac{2(D^2+\bw)}{\eps} \right). 
\end{equation*}
Now, for each invididual process $Y_\theta$ is the i.i.d. average of
random variables of the form $\eps_i\log\sum_{j=1}^k\exp(\mu_j^\top
X+c_j)$. The log-sum-exp part is $D$-Lipschitz in $X$, so we can reuse
the analysis done precedingly (in the Bobkov-Gotze part) to get that
$\log\sum_{j=1}^k\exp(\mu_j^\top X+c_j)$ is
$D^2(D^2+1)$-sub-Gaussian. Further, its expectation is bounded as (for
$X\sim p=\sum v_i\normal(\nu_i,I_d)$)
\begin{eqnarray*}
  && \E_{X\sim p}\left[\log\sum_{j=1}^k\exp(\mu_j^\top X + c_j)\right] =
     \sum_{i=1}^k
     v_i\E_{X\sim\normal(\nu_i,I_d)}\left[\log\sum_{j=1}^k\exp(\mu_j^\top
     X + c_j)\right] \\
  &\le& \sum_{i=1}^k
        v_i\log\sum_{j=1}^k\E_{X\sim\normal(\nu_i,I_d)}[\exp(\mu_j^\top
        X+c_j)] \le \sum_{i=1}^k v_i\log\sum_{j=1}^k
        \exp(\mu_j^\top\nu_i + \ltwo{\mu_j}^2/2+c_j) \\
  &\le& \log k + (2D^2+\bw).
\end{eqnarray*}
This shows that the term $\eps_i\log\sum_{j=1}^k\exp(\mu_j^\top
X+c_j)$ is $(\log k + D^2 + \bw)^2 + D^2(D^2+1)$-sub-Gaussian, and
thus we have by sub-Gaussian maxima bounds that
\begin{eqnarray}
  && \E\left[\max_{\theta\in\mc{N}(\Theta,\rho,\eps)} |Y_\theta|\right]
     \le C\sqrt{\frac{(\log k + D^2 + \bw)^2 + D^2(D^2+1)}{n}\cdot
     \log N(\Theta,\rho,\eps)} \nonumber \\
  &\le& C\sqrt{\frac{\log k + D^2 + \bw}{n} \cdot
        kd\log\frac{D^2+\bw}{\eps}}. \label{equation:mog-finite-maxima}
\end{eqnarray}
By the 1-step discretization bound and
combining~\cref{equation:mog-discretization-error}
and~\cref{equation:mog-finite-maxima}, we get
\begin{eqnarray*}
  && \E\left[\sup_{\theta\in\Theta}|Y_\theta|\right] \le \E\left[
     \sup_{\theta,\theta'\in\Theta,\rho(\theta,\theta')\le\eps}
  |Y_\theta-Y_{\theta'}|  \right] +
  \E\left[\max_{\theta\in\mc{N}(\Theta,\rho,\eps)} |Y_\theta|\right]
  \\
  &\le& Ck(D+\sqrt{d})\eps + C\sqrt{\frac{\log k + D^2 + \bw}{n} \cdot
        kd\log\frac{D^2+\bw}{\eps}}.
\end{eqnarray*}
Choosing $\eps=c/n$ for sufficiently small $c$ (depending on
$D^2,\bw$) gives that
\begin{equation*}
  \E\left[\sup_{\theta\in\Theta}|Y_\theta|\right] \le
  C\sqrt{\frac{kd(\log k+D^2+\bw)\log n}{n}}
\end{equation*}

%% file: cleantex-new/proof-generator.tex
\section{Proofs for Section~\ref{section:generator}}
\label{appendix:proof-generator}
\subsection{Bounding KL by Wasserstein}
The following theorem gives conditions on which the KL divergence can
be lower bounded by the Wasserstein 1/2 distance. For a reference
see Section 4.1 and 4.4 in \citet{VanHandel14}.
\begin{theorem}[Lower bound KL by Wasserstein]
  \label{theorem:kl-ge-wasserstein}
  Let $p$ be any distribution on $\R^d$ and $X_i\simiid p$ be the
  i.i.d. samples from $p$.
  \begin{enumerate}[(a)]
  \item (Bobkov-Gotze) If $f(X_1)$ is $\sigma^2$-sub-Gaussian for any
    $1$-Lipschitz $f:\R^d\to\R$, then
    \begin{equation*}
      W_1(p, q)^2 \le 2\sigma^2\dkl(p\|q)~~~\textrm{for all}~q.
    \end{equation*}
  \item (Gozlan) If $f(X_1,\dots,X_n)$ is $\sigma^2$-sub-Gaussian for
    any $1$-Lipschitz $f:(\R^d)^n\to \R$, then
    \begin{equation*}
      W_2(p, q)^2 \le 2\sigma^2\dkl(p\|q)~~~\textrm{for all}~q.
    \end{equation*}
  \end{enumerate}
\end{theorem}

\begin{theorem}[Upper bounding $f$-contrast by Wasserstein]
  \label{theorem:df-le-wasserstein}
  Let $p,q$ be two distributions on $\R^d$ with positive densities and
  denote their probability measures by $P,Q$. Let $f:\R^d\to\R$ be a
  function.
  \begin{enumerate}[(a)]
  \item ($W_1$ bound) Suppose $f$ is $L$-Lipschitz, then $\E_p[f(X)] -
    \E_q[f(X)] \le L\cdot W_1(p, q)$.
  \item (Truncated $W_1$ bound) Let $D>0$ be any diameter of
    interest. Suppose for any $\wt{p}\in\set{p,q}$ we have
    \begin{enumerate}[(i)]
    \item $f$ is $L(D)$-Lipschitz in the ball of
      radius $D$;
    \item We have $\E_{\wt{p}}[f^2(X)] \le M$;
    \item We have $\wt{P}(\norm{X}\ge D) \le \ptail(D)$,
    \end{enumerate}
    then we have
    \begin{equation*}
      \E_p[f(X)] - \E_q[f(X)] \le L(D) \cdot W_1(p, q) +
      4\sqrt{M\ptail(D)}.
    \end{equation*}
  \item ($W_2$ bound) Suppose
    $\norm{\grad f(x)}\le c_1\norm{x} + c_2$ for all $x\in\R^d$, then
    we have
    \begin{equation*}
      \E_p[f(X)] - \E_q[f(X)] \le \left(  
        \frac{c_1}{2}\sqrt{\E_p[\norm{X}^2]} +
        \frac{c_1}{2}\sqrt{\E_q[\norm{X}^2]} + c_2\right) \cdot
      W_2(p, q). 
    \end{equation*}
  \end{enumerate}
\end{theorem}

\subsubsection{Proof of Theorem~\ref{theorem:df-le-wasserstein}}
\begin{proof}
  \begin{enumerate}[(a)]
  \item This follows from the dual formulation of $W_1$.
  \item We do a truncation argument. We have
    \begin{align*}
      \E_p[f(X)] - \E_q[f(X)] =
      &
        \underbrace{\E_p[f(X)\indic{\norm{X}\le D}] -
        \E_q[f(X)\indic{\norm{X}\le D}]}_{\rm I} \\
      + &
         \underbrace{\E_p[f(X)\indic{\norm{X}> D}]
        - \E_q[f(X)\indic{\norm{X}> D}]}_{\rm II}.
    \end{align*}
    Term II has the followng bound by Cauchy-Schwarz:
    \begin{equation*}
      {\rm II} \le \sqrt{\E_p[f^2(X)] \cdot P(\norm{X}> D)} +
      \sqrt{\E_q[f^2(X)] \cdot Q(\norm{X}> D)} \le
      2\sqrt{M\ptail(D)}.
    \end{equation*}
    We now deal with term I. By definition of the Wasserstein
    distance, there exists a coupling $(X,Y)\sim\pi$ such that $X\sim
    P$, $Y\sim Q$, and $\E_\pi[\norm{X-Y}]=W_1(p,q)$. On this
    coupling, we have
    \begin{eqnarray*}
      {\rm I}
      &=& \E_\pi[f(X)\indic{\norm{X}\le D} -
          f(Y)\indic{\norm{Y}\le D}] \\ 
      &=& \E_\pi[(f(X) - f(Y))\indic{\norm{X}\le D,\norm{Y}\le D}] \\
      && + \E_\pi[f(X)\indic{\norm{X}\le D,\norm{Y}>D}]
         - \E_\pi[f(Y)\indic{\norm{Y}\le D,\norm{X}>D}] \\
      &\stackrel{(i)}{\le}& L(D)\E_\pi[\norm{X-Y}\indic{\norm{X}\le
                            D,\norm{Y}\le D}] +
                            \E_\pi[|f(X)|\indic{\norm{Y}\ge D}] \\
                            && + 
                            \E_\pi[|f(Y)|\indic{\norm{X}\ge D}] \\
      &\stackrel{(ii)}{\le}& L(D)\E_\pi[\norm{X-Y}] +
                             \sqrt{\E_\pi[f^2(X)]\cdot \pi(\norm{Y}\ge 
                             D)} \\ 
                         & & + \sqrt{\E_\pi[f^2(Y)] \cdot
                             \pi(\norm{X}\ge D)} \\
      &=& L(D)\cdot W_1(p, q) +
          \sqrt{\E_p[f^2(X)]\cdot Q(\norm{X}\ge 
          D)} + \sqrt{\E_q[f^2(X)] \cdot P(\norm{X}\ge D)} \\
      &\le& L(D)\cdot W_1(p, q) + 2\sqrt{M\ptail(D)}.
    \end{eqnarray*}
    Above, inequality (i) used the Lipschitzness of $f$ in the
    $D$-ball, and (ii) used Cauchy-Schwarz.
    Putting terms I and II together we get
    \begin{equation*}
      \E_p[f(X)] - \E_q[f(X)] \le L(D) \cdot W_1(p, q) +
      4\sqrt{M\ptail(D)}. 
    \end{equation*}
  \item This part is a straightforward extension of~\citep[Proposition
    1]{PolyanskiyWu16}. For completeness we present the proof
    here. For any $x,y\in\R^d$ we have
    {\small 
      \begin{eqnarray*}
        && |f(x) - f(y)| = \left| \int_0^1 \<\grad f(tx+(1-t)y), x-y\>
           dt\right| \le \int_0^1 \norm{\grad f(tx+(1-t)y)} \norm{x-y}dt
        \\
        &\le& \int_0^1 (c_1t\norm{x} + c_1(1-t)\norm{y} + c_2)
              \norm{x-y}dt = 
              (c_2 + c_1\norm{x}/2 + c_1\norm{y}/2)\norm{x-y}.
      \end{eqnarray*}
    }
  By definition of the $W_2$ distance, there exists a coupling
  $(X,Y)\sim\pi$ such that $X\sim P$, $Y\sim Q$, and
  $\E[\norm{X-Y}^2]=W_2^2(p, q)$. On this coupling, taking
  expectation of the above bound, we get
    \begin{eqnarray*}
      && \E_\pi[|f(X) - f(Y)|] \le c_2\E_\pi[\norm{X-Y}] +
         \frac{c_1}{2}\left(  
         \E_\pi[\norm{X}\norm{X-Y}] + \E_\pi[\norm{Y}\norm{X-Y}]
         \right) \\ 
      &\le& c_2\sqrt{\E_\pi[\norm{X-Y}^2]} +
            \frac{c_1}{2}\left(\sqrt{\E_\pi[\norm{X}^2] \cdot
            \E_\pi[\norm{X-Y}^2]} 
            + \sqrt{\E_\pi[\norm{Y}^2] \cdot
            \E_\pi[\norm{X-Y}^2]}\right) \\ 
      &=& \left(c_2 + \frac{c_1}{2}\sqrt{\E_p[\norm{X}^2]} +
          \frac{c_1}{2}\sqrt{\E_q[\norm{X}^2]} \right) \cdot W_2(p,
          q).
    \end{eqnarray*}
    Finally, the triangle inequality gives
    \begin{equation*}
      \E_p[f(X)] - \E_q[f(X)] = \E_\pi[f(X) - f(Y)] \le \E_\pi[|f(X) - 
      f(Y)|],
    \end{equation*}
    so the left hand side is also bounded by the preceding quantity.
  \end{enumerate}
\end{proof}

\subsection{Proof of Lemma~\ref{lemma:logp-neural-network}}
\label{appendix:proof-logp-neural-network}
It is straightforward to see that the inverse of $x=G_\theta(z)$ can
be computed as
\begin{equation}
  \label{equation:inverse-network}
  z =
  W_1^{-1}(\sigma^{-1}(W_2^{-1}\sigma^{-1}(\cdots\sigma^{-1}(W_{\ell}^{-1}(x-b_\ell))
  \cdots) - b_2) - b_1).
\end{equation}
So $G_\theta^{-1}$ is also a $\ell$-layer feedforward net with
activation $\sigma^{-1}$.

We now consider the problem of representing $\log p_\theta(x)$ by a
neural network. Let $\phi_{\gamma}$ be the density of
$Z\sim\normal(0,\diag(\gamma^2))$.
Recall that the log density has the formula
\begin{equation*}
  p_\theta(x) = \log\phi_{\gamma}\left((G_\theta^{-1}(x)\right) + 
  \log\left|\det\frac{\partial G_\theta^{-1}(x)}{\partial x}\right|.
\end{equation*}

First consider the inverse network that implements
$G_\theta^{-1}$. By~\cref{equation:inverse-network}, this network has
$\ell$ layers ($\ell-1$ hidden layers), $d^2+d$ parameters in each
layer, and $\sigma^{-1}$ as the activation function. Now, as
$\log\phi_{\gamma}$ has the form
$\log\phi_{\gamma}(z)=a(\gamma) - \sum_{i}z_i^2/(2\gamma_i^2)$, we can
add one more layer on top of $z$ with the square activation and the
inner product with $-\gamma^{-2}/2$ to get this term.

Second, we show that by adding some branches upon this
network, we can also compute the log determinant of the
Jacobian. Define $h_\ell=W_\ell^{-1}(x-b_\ell)$ and backward recursively
$h_{k-1}=W_{k-1}^{-1}(\sigma^{-1}(h_k)-b_{k-1})$ (so that $z=h_1$),
we have
\begin{equation*}
  \frac{\partial G_\theta^{-1}(x)}{\partial x} =
  W_1^{-1}\diag(\sigma^{-1'}(h_2))W_{2}^{-1}\cdots
  W_{\ell-1}^{-1}\diag(\sigma^{-1'}(h_\ell))W_\ell^{-1}.
\end{equation*}
Taking the log determinant gives
\begin{equation*}
  \log\left|\det\frac{\partial G_\theta^{-1}(x)}{\partial x}\right|
  = C + \sum_{k=2}^{\ell} \<\ones, \log\sigma^{-1'}(h_k)\>.
\end{equation*}
As $(h_\ell,\dots,h_2)$ are exactly the (pre-activation) hidden
layers of the inverse network, we can add one branch from each layer,
pass it through the $\log\sigma^{-1'}$ activation, and take the inner
product with $\ones$.

Finally, by adding up the output of the density branch and the log
determinant branch, we get a neural network that computes $\log
p_\theta(x)$ with no more than $\ell+1$ layers and $O(\ell d^2)$
parameters, and choice of activations within $\{\sigma^{-1},
\log\sigma^{-1'},(\cdot)^2\}$.

\subsection{Proof of Theorem~\ref{theorem:invertible-generator}}
We state a similar restricted approximability bound here in terms of
the $W_2$ distance, which we also prove.
\begin{equation*}
  W_2(p,q)^2 \lesssim \dkl(p\|q)+ \dkl(q\|p) \leq\df(p,q)  \lesssim
  \frac{\sqrt{d}}{\delta^2}\cdot W_2(p,q).
\end{equation*}
The theorem follows by
combining the following three lemmas, which we show in sequel.

\label{section:proof-invertible-generator}
\begin{lemma}[Lower bound]
  \label{lemma:invertible-generator-lower-bound}
  There exists a constant $c=c(\rw,\rb,\ell)>0$ such that for any
  $\theta_1,\theta_2\in\Theta$, we have
  \begin{equation*}
    \df(p_{\theta_1}, p_{\theta_2}) \ge c\cdot W_2(p_{\theta_1},
    p_{\theta_2})^2 \ge c\cdot W_1(p_{\theta_1}, p_{\theta_2})^2.
  \end{equation*}
\end{lemma}

\begin{lemma}[Upper bound]
  \label{lemma:invertible-generator-upper-bound}
  There exists constants $C_i=C_i(\rw,\rb,\ell)>0$, $i=1,2$ such that
  for any $\theta_1,\theta_2\in\Theta$, we have
  \begin{enumerate}[(1)]
  \item ($W_1$ bound) $\df(p_{\theta_1}, p_{\theta_2}) \le
    \frac{C_1\sqrt{d}}{\delta^2} \cdot \left(W_1(p_{\theta_1}, 
      p_{\theta_2}) + \sqrt{d}\exp(-10d)\right)$.
  \item ($W_2$ bound) $\df(p_{\theta_1}, p_{\theta_2}) \le
    \frac{C_2\sqrt{d}}{\delta^2} \cdot W_2(p_{\theta_1},
    p_{\theta_2})$.
  \end{enumerate}
\end{lemma}

\begin{lemma}[Generalization error]
  \label{lemma:invertible-generator-generalization}
  Consider $n$ samples $X_i\simiid p_{\theta^\star}$ for some
  $\theta^\star\in\Theta$. There exists a constant $C=C(\rw, \rb,
  \ell)>0$ such that when $n\ge C\max\set{d, \delta^{-8}\log n}$, we
  have
  \begin{equation*}
    R_n(\mc{F}, p_{\theta^\star}) \le \sqrt{\frac{Cd^4\log
        n}{\delta^4 n}}.
  \end{equation*}
\end{lemma}

\subsection{Proof of
  Lemma~\ref{lemma:invertible-generator-lower-bound}}
We show that $p_\theta$ satisfies the Gozlan condition for any
$\theta\in\Theta$ and apply
Theorem~\ref{theorem:kl-ge-wasserstein}. Let $X_i\simiid p_\theta$ for 
$i\in[n]$. By definition, we can write
\begin{equation*}
  X_i = G_\theta(Z_i),~~~Z_i\simiid \normal(0, \diag(\gamma^2)).
\end{equation*}
Let $\wt{Z}_i=(z_i/\gamma_i)_{i=1}^d$ and
$\wt{G}_\theta(\wt{z})=G_\theta((\gamma_i\wt{z}_i)_{i=1}^d)$,
then we have $G_\theta(z)=\wt{G}_\theta(\wt{z})$ and that $\wt{Z}_i$
are i.i.d. standard Gaussian. Further, suppose $G_\theta$ is
$L$-Lipschitz, then for all $z_1,z_2\in\R^d$ we have
\begin{equation*}
  \left|\wt{G}_\theta(\wt{z}_1) - \wt{G}_\theta(\wt{z}_2) \right| =
  \left|G_\theta(z_1) - G_\theta(z_2)\right| \le L\ltwo{z_1 - z_2} \le
  L\ltwo{\wt{z}_1 - \wt{z}_2},
\end{equation*}
the last inequality following from $\gamma_i\le 1$. Therefore
$\wt{G}_\theta$ is also $L$-Lipschitz.

Now, for any 1-Lipschitz $f:(\R^d)^n\to \R$, we have
\begin{eqnarray*}
  && \left| f(\wt{G}_{\theta}(\wt{z}_1),\dots,\wt{G}_{\theta}(\wt{z}_n)) -
     f(\wt{G}_{\theta}(\wt{z}_1'),\dots,\wt{G}_{\theta}(\wt{z}_n')) \right|
     \le \left(\sum_{i=1}^n \ltwo{\wt{G}_\theta(\wt{z}_i) -
     \wt{G}_\theta(\wt{z}_i')}^2 \right)^{1/2} \\
  &\le& L\cdot \left(\sum_{i=1}^n \ltwo{\wt{z}_i - \wt{z}_i'}^2
        \right)^{1/2} = L\ltwo{\wt{z}_{1:n} - \wt{z'}_{1:n}}.
\end{eqnarray*}
Therefore the mapping $(\wt{z}_1,\dots,\wt{z}_n)\to
f(\wt{G}_\theta(\wt{z}_1),\dots,\wt{G}_\theta(\wt{z}_n))$ is
$L$-Lipschitz. Hence by Lemma~\ref{lemma:gaussian-concentration}, the
random variable
\begin{equation*}
  f(X_1,\dots,X_n) =
  f(\wt{G}_\theta(\wt{Z}_1),\dots,\wt{G}_\theta(\wt{Z}_n))
\end{equation*}
is $L^2$-sub-Gaussian, and thus the Gozlan condition is satisfied with
$\sigma^2=L^2$. By definition of the network $G_\theta$ we have
\begin{equation*}
  L \le \prod_{k=1}^\ell\opnorm{W_k} \cdot \csigma^{\ell-1} \le
  \rw^\ell \csigma^{\ell-1} = C(\rw, \ell).
\end{equation*}
Now, for any $\theta_1,\theta_2\in\Theta$, we can apply
Theorem~\ref{theorem:kl-ge-wasserstein}(b) and get
\begin{equation*}
  \dkl(p_{\theta_1}\|p_{\theta_2}) \ge \frac{1}{2C^2}W_2(p_{\theta_1},
  p_{\theta_2})^2,
\end{equation*}
and the same holds with $p_{\theta_1}$ and $p_{\theta_2}$ swapped.
As $\log p_{\theta_1}-\log p_{\theta_2}\in\cF$, by Lemma~\ref{thm:logp},
we obtain
\begin{equation*}
  \df(p_{\theta_1},p_{\theta_2}) \ge \dkl(p_{\theta_1}\|p_{\theta_2})
  + \dkl(p_{\theta_2}\|p_{\theta_1}) \ge \frac{1}{C^2}W_2(p_{\theta_1},
  p_{\theta_2})^2 \ge \frac{1}{C^2}W_1(p_{\theta_1}, p_{\theta_2})^2,
\end{equation*}
The last bound following from the fact that $W_2\ge W_1$.

\subsection{Proof of
  Lemma~\ref{lemma:invertible-generator-upper-bound}}
We are going to upper bound $\df$ by the Wasserstein distances through
Theorem~\ref{theorem:df-le-wasserstein}. Fix
$\theta_1,\theta_2\in\Theta$. By definition of $\mc{F}$, it 
suffices to upper bound the Lipschitzness of $\log p_\theta(x)$ for
all $\theta\in\Theta$. Recall that
\begin{equation*}
  \log p_{\theta}(x) =
  \underbrace{\frac{1}{2}\<h_\ell, \diag(\gamma^{-2})h_\ell\>}_{\rm I} +
  \underbrace{\sum_{k=1}^{\ell-1} \<\ones_d,
    \log\sigma^{-1'}(h_k)\>}_{\rm II} + C(\theta),
\end{equation*}
where $h_1,\dots,h_\ell(=z)$ are the hidden-layers of the inverse
network $z=G_\theta^{-1}(x)$, and $C(\theta)$ is a constant that does
not depend on $x$.

We first show the $W_2$ bound. Clearly $\log p_\theta(x)$ is
differentiable in $x$. As $\theta\in\Theta$ has norm bounds, each
layer $h_k$ is $C(\rw,\rb,k)$-Lipschitz in $x$, so term II is
altogether $\sqrt{d}\bsigma\sum_{k=1}^{\ell-1}C(\rw,\rb,k) =
C(\rw,\rb,\ell)\sqrt{d}$-Lipschitz in $x$. For term I, note that
$h_\ell$ is $C$-Lipschitz in $x$, so we have
\begin{equation*}
  \ltwo{\grad_x {\rm I}} \le
  \frac{1}{\min_{i}\gamma_i^2}\ltwo{h_\ell(x)}\ltwo{\grad_x h_\ell(x)} \le
  \frac{1}{\delta^2}(C\ltwo{x} + h_\ell(0))\cdot C \le
  \frac{C}{\delta^2}(1+\ltwo{x}).
\end{equation*}
Putting together the two terms gives
\begin{equation}
  \label{equation:grad-logp}
  \ltwo{\grad \log p_\theta(x)} \le \frac{C}{\delta^2}(1+\ltwo{x}) +
  \frac{C}{\delta^2}\sqrt{d} \le \frac{C}{\delta^2}(\ltwo{x} + \sqrt{d}).
\end{equation}
Further, under either $p_{\theta_1}$ or $p_{\theta_2}$ (for example
$p_{\theta_1}$), we have
\begin{equation*}
  \E_{p_{\theta_1}}[\ltwo{X}^2] \le \E[\ltwo{G_{\theta_1}(Z)}^2] \le
  C(\rw,\rb,\ell)\E[(\ltwo{Z}+1)^2] \le Cd.
\end{equation*}
Therefore we can apply Theorem~\ref{theorem:df-le-wasserstein}(c) and
get
\begin{equation*}
  \E_{p_{\theta_1}}[\log p_{\theta}(x)] - \E_{p_{\theta_2}}[\log
  p_{\theta}(x)] \le \frac{C}{\delta^2}\left(2\sqrt{Cd} +
    \sqrt{d}\right) W_2(p_{\theta_1}, p_{\theta_2}) \le
  \frac{C\sqrt{d}}{\delta^2}W_2(p_{\theta_1}, p_{\theta_2}). 
\end{equation*}

We now turn to the $W_1$ bound. The bound~\cref{equation:grad-logp}
already implies that for $\ltwo{X}\le D$,
\begin{equation*}
  \ltwo{\grad\log p_\theta(x)} \le \frac{C}{\delta^2}(D+\sqrt{d}).
\end{equation*}
Choosing $D=K\sqrt{d}$, for a sufficiently large constant $K$, by the
bound $\ltwo{X}\le C(\ltwo{Z}+1)$ we have the tail bound
\begin{equation*}
  \P(\ltwo{X} \ge D) \le \exp(-20d).
\end{equation*}
On the other hand by the bound $|\log p_\theta(x)|\le
C((\ltwo{x}+1)^2/\delta^2+\sqrt{d}(\ltwo{x}+1))$ we get under either
$p_{\theta_1}$ or $p_{\theta_2}$ (for example $p_{\theta_1}$) we have
\begin{equation*}
  \E_{p_{\theta_1}}\left[(\log p_\theta(X))^2\right] \le
  \frac{C}{\delta^4}\E\left[\left(\ltwo{X}^2 +
    \sqrt{d}(\ltwo{X}+1)\right)^2\right] \le \frac{Cd^2}{\delta^4}.
\end{equation*}
Thus we can substitute $D=K\sqrt{d}$, $L(D)=C(1+K)\sqrt{d}/\delta^2$,
$M=Cd^2/\delta^4$, and $\ptail(D)=\exp(-2d)$ into
Theorem~\ref{theorem:df-le-wasserstein}(b) and get
\begin{eqnarray*}
  && \E_{p_{\theta_1}}[\log p_{\theta}(x)] - \E_{p_{\theta_2}}[\log
     p_{\theta}(x)] \le
     \frac{C(1+K)\sqrt{d}}{\delta^2}W_1(p_{\theta_1},p_{\theta_2}) +
     4\sqrt{\frac{Cd^2}{\delta^4}\exp(-20d)} \\
  &\le& \frac{C\sqrt{d}}{\delta^2}\left(
        W_1(p_{\theta_1},p_{\theta_2}) + 
        \sqrt{d}\exp(-10d) \right).
\end{eqnarray*}

\subsection{Proof of
  Lemma~\ref{lemma:invertible-generator-generalization}}
For any log-density neural network $F_\theta(x)=\log
p_\theta(x)$, reparametrize so that $(W_i, b_i)$ represent the weights
and the biases of the inverse network
$z=G_\theta^{-1}(x)$. By~\cref{equation:inverse-network}, this has
the form
\begin{equation*}
  (W_i, b_i) \longleftarrow (W_{\ell-i+1}^{-1},
  -W_{\ell-i+1}^{-1}b_{\ell-i+1}),~~\forall i\in[\ell].
\end{equation*}
Consequently the reparametrized $\theta=(W_i,b_i)_{i\in[\ell]}$
belongs to the (overloading $\Theta$)
\begin{equation}
  \label{equation:Theta-reparametrized}
  \Theta = \set{\theta = (W_i,
    b_i)_{i=1}^\ell:~\max\set{\opnorm{W_i},
      \opnorm{W_i^{-1}}} \le \rw,~\ltwo{b_i}\le \rw\rb,~\forall
    i\in[\ell]}.
\end{equation}

As $\mc{F}=\set{F_{\theta_1} -
  F_{\theta_2}:\theta_1,\theta_2\in\Theta}$, the Rademacher complexity 
of $\mc{F}$ is at most two times the quantity
\begin{equation*}
  R_n \defeq  R_n(\set{F_\theta:\theta\in\Theta}, p_{\theta^\star}) =
  \left[\sup_{\theta\in\Theta} \left|\frac{1}{n}\sum_{i=1}^n \eps_i
    F_\theta(X_i)\right| \right],
\end{equation*}
We do one additional re-parametrization. Note that the log-density
network $F_\theta(x)=\log p_\theta(x)$ has the form
\begin{equation}
  \label{equation:log-density-network}
  F_\theta(x) = \phi_{\gamma}(G_\theta^{-1}(x)) +
  \log\left|\det\frac{\partial G_\theta^{-1}(x)}{\partial x}\right| +
  K(\theta) = \frac{1}{2}\<h_\ell, \diag(\gamma^{-2})h_\ell\> +
  \log\left|\det\frac{\partial
      G_\theta^{-1}(x)}{\partial x}\right| + K(\theta).
\end{equation}
The constant $C(\theta)$ is the sum of the normalizing constant for
Gaussian density (which is the same across all $\theta$, and as we are
taking subtractions of two $\log p_\theta$, we can ignore this) and
the sum of $\log\det(W_i)$, which is upper bounded by $d\ell\rw$. We
can additionally create a parameter $K=K(\theta)\in[0, d\ell\rw]$ for
this term and let $\theta\leftarrow(\theta, K)$.

For any (reparametrized)
$\theta, \theta'\in\Theta$, define the metric
\begin{equation*}
  \rho(\theta, \theta') = \max\set{\opnorm{W_i - W'_i},
    \ltwo{b_i - b'_i},|K-K'|:i\in[d]}.
\end{equation*}
Then we have, letting $Y_\theta=\frac{1}{n}\sum_{i=1}^n \eps_i
F_\theta(X_i)$ denote the Rademacher process, the one-step
discretization bound~\citep[Section 5]{Wainwright18}. 
\begin{equation}
  \label{equation:discretization-bound}
  R_n \le
  \E\left[\sup_{\theta,\theta'\in\Theta,\rho(\theta,\theta')\le
      \eps} |Y_\theta - Y_\theta'| \right] +
  \E\left[\sup_{\theta_i\in\mc{N}(\Theta, \rho, \eps)}
    |Y_{\theta_i}|\right].
\end{equation}
We deal with the two terms separately in the following two
lemmas. 

\begin{lemma}[Discretization error]
  \label{lemma:ig-discretization-error}
  There exists a constant $C=C(\rw,\rb,\ell)$ such that, for all
  $\theta,\theta'\in\Theta$ such that
  $\rho(\theta,\theta')\le \eps$, we have
  \begin{equation*}
    |Y_\theta - Y_{\theta'}| \le
    \frac{C}{\delta^2}\frac{1}{n}\sum_{i=1}^n 
    \left(\sqrt{d}(1+\ltwo{X_i}) + \ltwo{X_i}^2\right) \cdot \eps.
  \end{equation*}
\end{lemma}

\begin{lemma}[Expected max over a finite set]
  \label{lemma:ig-finite-max}
  There exists constants $\lambda_0,C$ (depending on $\rw,\rb,\ell$
  but not $d,\delta$) such that for all $\lambda \le
  \lambda_0\delta^2n$,
  \begin{equation*}
     \E\left[\max_{\theta_i\in\mc{N}(\Theta,\rho,\eps)}
       |Y_\theta|\right] \le
     \frac{Cd^2\log\frac{\max\set{\rw,\rb}}{\eps}}{\lambda} +
     \frac{Cd^2\lambda}{\delta^4 n}.
  \end{equation*}
\end{lemma}

Substituting the above two Lemmas into the
bound~\cref{equation:discretization-bound}, we get that for all
$\eps\le \min\set{\rw,\rb}$ and $\lambda\le\lambda_0\delta^2n$,
\begin{equation*}
  R_n \le
  \underbrace{\frac{C}{\delta^2}\E[\sqrt{d}(1+\ltwo{X})+\ltwo{X}^2]
    \cdot \eps}_{\rm I} +
  \underbrace{\frac{Cd^2\log\frac{\max\set{\rw,\rb}}{\eps}}{\lambda} + 
    \frac{Cd^2\lambda}{\delta^4 n}}_{\rm II}.
\end{equation*}
As $X=G_{\theta^\star}(Z)$, we have $\ltwo{X} \le
C\ltwo{Z}+\ltwo{G_{\theta^\star}(0)} \le C(\ltwo{Z}+1)$ for
some constant $C>0$. As $\E[\ltwo{Z}^2]=k+\delta^2(d-k)\le d$, we
have $\E[\sqrt{d}(1+\ltwo{X})+\ltwo{X}^2]\le Cd$ for some constant
$C$, giving that ${\rm I} \le \frac{Cd}{\delta^2}\cdot \eps$. Choosing
$\eps=c/(dn)$ guarantees that ${\rm I}\le \frac{1}{\delta^2 n}$. For
this choice of $\eps$, term II has the form
\begin{equation*}
  {\rm II} \le \frac{Cd^2\log(nd\max\set{\rw,\rb})}{\lambda} +
  \frac{Cd^2\lambda}{\delta^4 n} \le \frac{Cd^2\log n}{\lambda} +
  \frac{Cd^2\lambda}{\delta^4n},
\end{equation*}
the last bound holding if $n\ge d$. Choosing
$\lambda=\sqrt{n\log n/\delta^4}$, which will be valid if $n/\log n\ge
\delta^{-8}\lambda_0^{-2}$, we get
\begin{equation*}
  {\rm II} \le Cd^2\sqrt{\frac{\log n}{\delta^4 n}} =
  C\sqrt{\frac{d^4\log n}{\delta^4 n}}.
\end{equation*}
This term dominates term I and is hence the order of the
generalization error.

\subsubsection{Proof of Lemma~\ref{lemma:ig-discretization-error}}
Fix $\theta,\theta'$ such that $\rho(\theta,\theta')\le \eps$.
As $Y_\theta$ is the empirical average over $n$ samples and
$|\eps_i|\le 1$, it suffices to show that for any $x\in\R^d$,
\begin{equation*}
  |F_\theta(x) - F_{\theta'}(x)| \le
  \frac{C}{\delta^2}\left(\sqrt{d}(1 + \ltwo{x}) + 
  \ltwo{x}^2\right)\cdot\eps.
\end{equation*}
For the inverse network $G_\theta^{-1}(x)$, let $h_k(x)\in\R^d$ denote
the $k$-th hidden layer:
\begin{equation*}
  h_1(x) = \sigma(W_1x + b_1),~\cdots,~h_{\ell-1}(x) =
  \sigma(W_{\ell-1}h_{\ell-2}(x)+b_{\ell-1}),~h_{\ell}(x) = W_\ell
  h_{\ell-1}(x) + b_\ell = G_\theta^{-1}(x).
\end{equation*}
Let $h'_k(x)$ denote the layers of $G_{\theta'}^{-1}(x)$
accordingly. Using this notation, we have
\begin{equation*}
  F_\theta(x) =
  \frac{1}{2}\<h_\ell, \diag(\gamma^{-2})h_\ell\> +
  \sum_{k=1}^{\ell-1} \<\ones_d, \log\sigma^{-1'}(h_k)\> + K.
\end{equation*}

\paragraph{Lipschitzness of hidden layers}
We claim that for all $k$, we have
\begin{equation}
  \label{equation:hl-norm-bound}
  \ltwo{h_k} \le (\rw\csigma)^k\ltwo{x} +
  \rb\sum_{j=1}^{k}(\rw\csigma)^j, 
\end{equation}
and consequently when $\rho(\theta,\theta')\le\eps$, we have
\begin{equation}
  \label{equation:hl-lip}
  \ltwo{h_k - h'_k} \le C(\rw, \rb,
  k)\eps(1+\ltwo{x}),~~~C(\rw, \rb,
  k)=O\left(\sum_{j=0}^{k-1}j(\rw\csigma)^{j}(1+\rb)\right).
\end{equation}
We induct on $k$ to show these two
bounds. For~\cref{equation:hl-norm-bound}, note that $h_0=\ltwo{x}$
and
\begin{equation*}
  \ltwo{h_k} = \ltwo{\sigma(W_kh_{k-1}+b_k)} \le
  \csigma(\rw\ltwo{h_{k-1}} + \rw\rb),
\end{equation*}
so an induction on $k$ shows the bound. For~\cref{equation:hl-lip},
note that
\begin{equation*}
  \ltwo{h_1 - h_1'} \le \opnorm{W_1 - W_1'}\ltwo{x} + \ltwo{b_1 -
    b_1'} \le \eps(1 + \ltwo{x}),
\end{equation*}
so the base case holds. Now, suppose the claim holds for the
$(k-1)$-th layer, then for the $k$-th layer we have
{\small 
\begin{eqnarray*}
  \ltwo{h_k - h_k'}
  &=& \ltwo{\sigma(W_kh_{k-1}+b_k) -
      \sigma(W_kh_{k-1}'+b_k')} \le
      \csigma\left(\ltwo{W_kh_{k-1}-W_k'h_{k-1}'} + \ltwo{b_k -
      b_k'}\right) \\
  &\le& \csigma\left(\eps + \opnorm{W_k}\ltwo{h_{k-1} - h_{k-1}'} +
        \opnorm{W_k - W_k'}\ltwo{h_{k-1}'}\right) \\
  &\le& \csigma\left(\eps + \rw C(\rw,\rb,k-1)\eps(1+\ltwo{x}) +
        \eps\left((\rw\csigma)^{k-1}\ltwo{x} +
        \rb\sum_{j=1}^{k-1}(\rw\csigma)^j\right)\right) \\
  &\le& \eps(1+\ltwo{x})\underbrace{\left(\csigma\rw C(\rw,\rb,k-1) +
    (1+\rb)\sum_{j=1}^{k-1}(\rw\csigma)^j \right)}_{C(\rw,\rb,k)},
\end{eqnarray*}
}`
verifying the result for layer $k$.

\paragraph{Dealing with $(\cdot)^2$ and $\log\sigma^{-1'}$}
For the $\log\sigma^{-1'}$ term, note that $|(\log\sigma^{-1'})'| =
|\sigma^{-1''}/\sigma^{-1'}|\le\bsigma$ by assumption. So we have the
Lipschitzness
\begin{eqnarray*}
  && \left| \sum_{k=1}^{\ell-1}\<\ones_d, \log\sigma^{-1'}(h_k) -
  \log\sigma^{-1'}(h_k')\> \right| \le
  \sqrt{d}\bsigma\sum_{k=1}^{\ell-1}\ltwo{h_k - h_k'} \\
  &\le& \sqrt{d}
        \underbrace{\bsigma\sum_{k=1}^{\ell-1}C(\rw,\rb,k)}_{C} 
        \cdot \eps(1+\ltwo{x}) = C\sqrt{d}(1+\ltwo{x}) \cdot \eps.
\end{eqnarray*}
For the quadratic term, let
$A_\gamma=\diag(\gamma^{-2})$ for shorthand. Using the
bound $(1/2)|\<u,Au\>-\<v,Av\>| \le \opnorm{A}(\ltwo{v}\ltwo{u-v} +
\ltwo{u-v}^2/2)$, we get
\begin{eqnarray*}
  && \frac{1}{2}\left|\<h_\ell, A_\gamma h_\ell\> - \<h_\ell', A_\gamma
     h_\ell'\>\right| \le \opnorm{A_\gamma}
     \left(\ltwo{h_\ell}\ltwo{h_\ell - h'_\ell} +
     \ltwo{h_\ell - h'_\ell}^2/2\right) \\
  &\le& \frac{1}{\delta^2} \left( C\cdot
        C(\rw,\rb,\ell)\eps(1+\ltwo{x})^2 +
        C(\rw,\rb,\ell)^2\eps^2(1+\ltwo{x})^2/2 \right) \le
        \frac{C}{\delta^2}(1+\ltwo{x})^2\cdot \eps.
\end{eqnarray*}

\paragraph{Putting together}
Combining the preceding two bounds and that $|K-K'|\le\eps$, we get
\begin{equation*}
  |F_\theta(x) - F_{\theta'}(x)| \le
  \left(\frac{C}{\delta^2}(1+\ltwo{x})^2 +
    C\sqrt{d}(1+\ltwo{x}) + 1\right)\eps \le
  \frac{C}{\delta^2}(\sqrt{d}(1+\ltwo{x})+\ltwo{x}^2)\cdot \eps.
\end{equation*}

\subsubsection{Proof of Lemma~\ref{lemma:ig-finite-max}}
\paragraph{Tail decay at a single $\theta$}
Fixing any $\theta\in\Theta$, we show that the random variable
\begin{eqnarray*}
  Y_\theta = \frac{1}{n}\sum_{i=1}^n \eps_iF_\theta(x_i) =
  \frac{1}{n}\sum_{i=1}^n \eps_i\left( \<h_\ell(x_i), A_\gamma
  h_\ell(x_i)\> + \sum_{k=1}^{\ell-1}\<\ones_d,
  \log\sigma^{-1'}(h_k(x_i))\> + K\right).
\end{eqnarray*}
is suitably sub-exponential. To do this, it suffices to look at a
single $x$ and then use rules for independent sums.

First, each $\<\ones_d, \log\sigma^{-1}(h_k(x))\>$ is sub-Gaussian,
with mean and sub-Gaussianity parameter
$O(Cd)$. Indeed, we have
\begin{equation*}
  \<\ones_d, \log\sigma^{-1}(h_k(x))\>
  = \<\ones_d, \log\sigma^{-1}(h_k(G_{\theta^\star}(z)))\>
  = \<\ones_d, \log\sigma^{-1}(h_k(\wt{G}_{\theta^\star}(\wt{z})))\>,
\end{equation*}
where $\wt{z} = [z_{1:k}, z_{(k+1):d}/\delta]$ is standard Gaussian.
Note that (1) Lipschitzness of $\wt{G}$ is bounded by that of $G$,
which is some $C(\rw,\rb,\ell)$, (2) all hidden-layers are
$C(\rw,\rb,\ell)$-Lipschitz (see~\cref{equation:hl-lip}), (3)
$v\mapsto \sum_{j=1}^d\log\sigma^{-1}(v_j)$ is
$\sqrt{d}\bsigma$-Lipschitz. Hence the above term is
a $C\sqrt{d}$-Lipschitz function of a standard Gaussian, so is
$Cd$-sub-Gaussian by Gaussian
concentration~\ref{lemma:gaussian-concentration}. To bound the 
mean, use the bound 
\begin{equation*}
 \E\left|\<\ones_d,
   \log\sigma^{-1}(h_k(\wt{G}_{\theta^\star}(\wt{z})))\>\right|
 \le \E[\sqrt{d}\bsigma C\ltwo{\wt{z}}] \le Cd.
\end{equation*}
As we have $\ell-1$ terms of this form, their sum is still
$Cd$-sub-Gaussian with a $O(Cd)$ mean (absorbing $\ell$ into $C$).

Second, the term $\<h_\ell, A_\gamma h_\ell\>$ is a quadratic function
of a sub-Gaussian random vector, hence is sub-exponential. Its mean is
bounded by $\E[\opnorm{A_\gamma}\ltwo{h_\ell}^2]\le Cd/\delta^2$. Its
sub-exponential parameter is $1/\delta^2$ times the sub-Gaussian
parameter of $h_\ell$, hence also $Cd/\delta^2$. In particular, there
exists a constant $\lambda_0>0$ such that for all
$\lambda\le\lambda_0\delta^2$,
\begin{equation*}
  \E[\exp(\lambda\<h_\ell, A_\gamma h_\ell\>)] \le \exp\left(
    \frac{Cd}{\delta^2}\lambda +
    \frac{C^2d^2\lambda^2}{\delta^4}\right).
\end{equation*}
(See for example~\citep{Vershynin10} for such results.) Also, the
parameter $K$ is upper bounded by $d\ell\rw=Cd$.

Putting together, multiplying by $\eps_i$ (which addes up the squared
mean onto the sub-Gaussianity / sub-exponentiality and multiplies
it by at most a constant)
and summing over $n$, we get that $Y_\theta$ is mean-zero
sub-exponential with the MGF bound 
\begin{eqnarray}
  && \E[\exp(\lambda Y_\theta)] = \left(\E\left[
    \exp\left(\frac{\lambda}{n}\eps_i\left( \<h_\ell(x_i), A_\gamma
        h_\ell(x_i)\> + \sum_{k=1}^{\ell-1}\<\ones_d,
        \log\sigma^{-1'}(h_k(x_i))\> \right) \right)\right]\right)^n
     \nonumber \\
  &\le& \exp\left(\frac{Cd^2\lambda^2}{\delta^4n}\right),~~~\forall
        \lambda\le\lambda_0\delta^2n.
        \label{equation:ytheta-mgf-bound} 
\end{eqnarray}

\paragraph{Bounding the expected maximum}
We use the standard covering argument to bound the expected
maximum. Recall that $\rho(\theta,
\theta')=\max\set{\opnorm{W_i-W_i'},\ltwo{b_i-b_i'},|K-K'|}$. Hence, the
covering number of $\Theta$ is bounded by the product of independent 
covering numbers, which further by the volume argument is
\begin{eqnarray*}
  && N(\Theta, \rho, \eps) \le \prod_{k=1}^\ell \left(1 +
     \frac{2\rw}{\eps}\right)^{d^2} \cdot \prod_{k=1}^\ell \left(1 +
     \frac{2\rb}{\eps} \right)^d\cdot
     \left(1+\frac{Cd\rw}{\eps}\right) \\
  &\le& \exp\left(C\left(\ell 
        d^2\log\frac{3\rw}{\eps} + \ell
        d\log\frac{3\rb}{\eps}\right)\right).
\end{eqnarray*}
Using Jensen's inequality and applying the
bound~\cref{equation:ytheta-mgf-bound}, we get that for any
$\lambda\le\lambda_0\delta^2n$,
\begin{eqnarray*}
  && \E\left[\max_{\theta_i\in\mc{N}(\Theta,\rho,\eps)} Y_\theta\right] \le
     \frac{1}{\lambda}\left( \log\E\left[\sum_{i=1}^{N(\Theta,\rho,\eps)}
     \exp(\lambda Y_{\theta_i})\right]\right) \le
     \frac{1}{\lambda}\left(\log N(\Theta,\rho,\eps) +
     \frac{Cd^2\lambda^2}{\delta^4 n}\right) \\
  &\le& \frac{Cd^2\log\frac{\max\set{\rw,\rb}}{\eps}}{\lambda} +
        \frac{Cd^2\lambda}{\delta^4 n}.
\end{eqnarray*}

%% file: cleantex-new/proof-additive.tex
\section{Proofs for Section~\ref{s:additive}}
\label{sec:proof:additive}

\subsection{Formal theorem statement}

Towards stating the theorem more quantitatively, we will need to specify a few quantities of the generator class that will be relevant for us. 

First, for notational simplicity, we override the definition of $p_{\theta}^{\beta}$ by a truncated version of the convolution of $p$ and a Gaussian distribution. Concretely, let $D_z = \{z: \|z\| \leq \sqrt{d} \log^2 d, z\in \R^k\}$ be a truncated region in the latent space (which contains an overwhelming large part of the probability mass), and the let $D_x = \{x: \exists z \in D_z, \|G(z) - x\|_2 \leq \beta \sqrt{d} \log^2 d\}$ be the image of $D_z$ under $G_{\theta}$. Recall that 
$$G_{\theta}(z) = \sigma\left(W_l \sigma\left(W_{l-1} \dots \sigma\left(W_1 z + b_1\right) + \dots b_{l-1}\right) + b_l\right).$$

Then, let $p^{\beta}_{\theta}(x)$ be the distribution obtained by adding Gaussian noise with variance $\beta^2$ to a sample from $G_{\theta}$, and truncates the distribution to a very high-probability region (both in the latent variable and observable domain.) Formally,  let $p^{\beta}_{\theta}$ be a distribution over $\mathbb{R}^d$, s.t. 
\begin{equation} p^{\beta}_{\theta}(x) \propto \int_{z \in D_z} e^{-\|z\|^2} e^{- \frac{\|G_{\theta}(z) - x\|^2}{\beta^2}}dz, \forall x \in D_x \label{eq:gen}\end{equation}

For notational convenience, denote by $f: \mathbb{R}^k \to \mathbb{R}$ the function $f(z) = -\|z\|^2 - \|G_{\theta}(z) - x\|^2/\beta^2$, and denote by $z^*$ a maximum of $f$.\\
Furthermore, whenever clear from the context, we will drop $\theta$ from $p_{\theta}$ and $G_{\theta}$.  

We introduce several regularity conditions for the family of generators $\mathcal{G}$: 
\begin{ass}\label{thm:ass}
	We assume the following bounds on the partial derivatives of $f$: we denote $S := \max_{z \in D_z:\|z - z*\| \leq \delta}\|\nabla^2 (\|G_{\theta}(z) - x\|^2)\|$, and $\lambda_{\min} := 
	\max_{z \in D_z:\|z - z*\| \leq \delta} \lambda_{\min}(\nabla^2 \|G_{\theta}(z) - x\|^2)$. \\
	Similarly, we denote $t(z):= k^3 \max_{|I| = 3} \frac{\partial \|G_{\theta}(z) - x\|^2}{\partial I}(z)$. 
	and $T = \max_{z: z \in D_z} |t(z)| $. \\
	We will denote by $R$ an upper bound on the quantity $\frac{1}{L^{l}_{\sigma}} \prod_{j=i}^L \sigma_{\min}(W_j) $ and by $L_G$ an upper bound on the quantity $L_G := 2^{l} L^l_{\sigma} \prod_{i=1}^l \frac{1}{\sigma_{\min}(W_i)}$.\\	Finally, we assume the inverse activation function is Lipschitz, namely $|\sigma^{-1}(x) - \sigma^{-1}(y)| \leq L_{\sigma} |x - y|$. 
\end{ass}
\noindent {\bf Note on asymptotic notation:} For notational convenience, in this section, $\lesssim,\gtrsim$, as well as the Big-Oh notation will hide dependencies on $R, L_G, S, T$ (in the theorem statements we intentionally emphasize the polynomial dependencies on $d$.)
The main theorem states that for certain $\cF$, $\tilde{d}_{\cF}$ approximates the Wasserstein distance. 
\begin{theorem}
  \label{theorem:additive-noise}
  Suppose the generator class $\cG$ satisfies the
  assumption~\ref{thm:ass} and let $\cF$ be the family of functions as
  defined in Theorem~\ref{thm:formal}. Then, we have that for every
  $p,q\in \cG$, 
  \begin{align}
    W_1(p,q) \lesssim \tilde{d}_{\cF}(p, q) \lesssim \mbox{poly(d)} \cdot W_1(p,q)^{1/6} + \exp(-d).
  \end{align}
	Furthermore, when $n \gtrsim \mbox{poly}(d)$ we have
$
    R_n(\mc{F}, \mathcal{G}) \lesssim  \mbox{poly(d)} \sqrt{\frac{\log
        n}{n}}.
$ Here $\lesssim$ hides dependencies on $R, L_G, S, $ and $T$.
\end{theorem}

The main ingredient in the proof will be the theorem that shows that there exists a parameterized family $\cF$ that can approximate the log density of $p^{\beta}$ for every $p\in \cG$. 
\begin{theorem}  \label{thm:formal} Let $\cG$ satisfy the assumptions in Assumption~\ref{thm:ass}. For $\beta = O(\mbox{poly}(1/d))$, there exists a family of neural networks $\cF$ of size $\mbox{poly}(\frac{1}{\beta}, d)$ such that  for every distribution $p\in \cG$, there exists $\NNN\in \cF$ satisfying: \\ 
	\\(1) $\NNN$ approximates $\log p$ for typical $x$: given input $x = G(z^*) + r$, for $\|r\| \leq 10 \beta \sqrt{d} \log d$, and $\|z^*\| \leq 10 \sqrt{d} \log d$ for $\beta = O(\mbox{poly}(1/d))$ it outputs $\NNN(x)$, s.t. 
$$|\NNN(x) - \log p^{\beta}(x) | = O_{\mbox{poly(d)}}(\beta \log(1/\beta)) + \exp(-d)$$
(2) $\NNN$ is globally an approximate lower bound of $p$: on any input $x$,  $\NNN$ outputs $\NNN(x) \leq \log p^{\beta}(x) + O_{\mbox{poly}(d)}(\beta \log(1/\beta)) + \exp(-d)$. \\
(3) $\NNN$ approximates the entropy in the sense that: the output $\NNN(x)$ satisfies $|\E_{p^{\beta}} \NNN(x) - H(p^{\beta})| = O_{\mbox{poly}(d)}(\beta \log(1/\beta)) + \exp(-d) $\\ 
Moreover, every function in $\cF$ has Lipschitz constant $O(\frac{1}{\beta^4} \mbox{poly}(d))$. 
\label{t:mainadditive}
\end{theorem}

The approach will be as follows: we will approximate $p^{\beta}(x)$ essentially by a variant of Laplace's method of integration, using the fact that
$$ p^{\beta}(x) = C \int_{z \in D_z} e^{-\|z\|^2 - \frac{\|G(z) - x\|^2}{\beta^2}} dz $$ 
for a normalization constant $C$ that can be calculated up to an exponentially small additive factor.  
When $x$ is typical (in case (1) of Theorem~\ref{thm:formal}), the integral will mostly be dominated by it's maximum value, which we will approximately calculate using a greedy ``inversion'' procedure. \\
When $x$ is a atypical, it turns out that the same procedure will give a lower bound as in (2).  

We are ready to prove Theorem  \ref{theorem:additive-noise}, assuming the correctness of Theorem~\ref{thm:formal}: 
\begin{proof}[Proof of Theorem \ref{theorem:additive-noise}]
	By Theorem~\ref{thm:formal}, we have that there exist neural networks $N_1, N_2\in \cF$ that approximate $\log p^\beta$ and $\log q^\beta$ respectively in the sense of bullet (1)-(3) in Theorem~\ref{thm:formal}. Thus we have that by bullet (2) for distribution $q^\beta$, and bullet (3) for distribution $q^\beta$, we have
	\begin{align}
	\Exp_{p^\beta}\left[N_1(x)- N_2(x)\right] \ge \Exp_{p^\beta}[\log p] - \Exp_{p^\beta}[\log q] - O(\beta \log 1/\beta)
	\end{align}
	Similarly, we have 	
		\begin{align}
	\Exp_{q^\beta}\left[N_2(x)- N_1(x)\right] \ge \Exp_{q^\beta}[\log q] - \Exp_{q^\beta}[\log p] - O(\beta \log 1/\beta)
	\end{align}
	Combining the equations above, setting $f = N_1(x) - N_2(x)$, we obtain that 
	\begin{align}\df(p^\beta, q^\beta)\ge \E_{p^\beta}\left[f\right] - \E_{q^\beta}\left[f\right]\ge \dkl(p^\beta\|q^\beta) + \dkl(q^\beta\| p^\beta) - O(\beta \log 1/\beta)
	\end{align}
	Therefore, by definition, and Bobkov-G{\"o}tze theorem ($\dkl(p^\beta, q^\beta)\gtrsim W_1(p^\beta, q^\beta)$)
	\begin{align}
W_1(p, q)  & \le W_1(p^\beta, q^\beta) + O(\beta)\lesssim (\dkl(p^\beta\|q^\beta) + \dkl(q^\beta\| p^\beta))^{1/2} + O(\beta) \nonumber\\
& \le (\df(p^\beta, q^\beta) + O(\beta\log(1/\beta))^{1/2} + O(\beta) \le O(\tilde{d}_{\cF}(p, q))
	\end{align}
Thus we prove the lower bound. 

Proceeding to the upper bound, notice that 
$W_{\mathcal{F}}(p^{\beta}, q^{\beta}) \lesssim \frac{1}{\beta^4} W_1(p^{\beta}, q^{\beta})$
since every function in $\mathcal{F}$ is $O(\mbox{poly}(d) \frac{1}{\beta^4})$-Lipschitz by Theorem~\ref{thm:formal}. We relate $ W_1(p^{\beta}, q^{\beta})$ to  $W_1(p, q)$, more precisely we prove: 
$ W_1(p^{\beta}, q^{\beta}) \leq (1+e^{-d}) W_1(p,q)$. Having this, we'd be done: namely, we simply set $\beta = W^{1/6}$ to get the necessary bound.

Proceeding to the claim, consider the optimal coupling $C$ of $p, q$, and consider the induced coupling $C_z$ on the latent variable $z$ in $p,q$. Then,  
$$W_1(p, q) = \int_{z \in \mathbb{R}^d} \|G(z) - G(z')\|_1 dC_z(z,z') \mbox{det}\left(\frac{\partial G_{\theta}(z)}{\partial z}\right) \mbox{det}\left(\frac{\partial G_{\theta}(z')}{\partial z'}\right)$$ 
Consider the coupling $\tilde{C}_z$ on the latent variables of $p^{\beta}, q^{\beta}$, specified as 
$\tilde{C}_z(z,z') = C(z,z') (1 - \Pr[z \notin D_z])^2$. The coupling $\tilde{C}$ of $p^{\beta},q^{\beta}$ specified by coupling $z$'s according to $\tilde{C}_z$ and the (truncated) Gaussian noise to be the same in $p^{\beta}, q^{\beta}$, we have that 
\begin{align*} W_1(p^{\beta}, q^{\beta}) &\leq \int_{z \in D_z} \|G(z) - G(z')\|_1 d\tilde{C}_z(z,z') \mbox{det}\left(\frac{\partial G_{\theta}(z)}{\partial z}\right) \mbox{det}\left(\frac{\partial G_{\theta}(z')}{\partial z'}\right) \\
&\leq \int_{z \in D_z} (1+e^{-d}) \|G(z) - G(z')\|_1 d C_z(z,z') \mbox{det}(\frac{\partial G_{\theta}(z)}{\partial z}) \mbox{det}\left(\frac{\partial G_{\theta}(z')}{\partial '}\right)  \\  
&\leq (1+e^{-d}) W_1(p,q) \end{align*}

The generalization claim follows completely analogously to  Lemma~\ref{lemma:invertible-generator-generalization}, using the Lipschitzness bound of the generators 
in Theorem~\ref{thm:formal}.

\end{proof}
The rest of the section is dedicated to the proof of Theorem~\ref{thm:formal}, which will be finally in Section~\ref{sec:t:mainadditive}.
\subsection{Tools and helper lemmas} 

First, we prove several helper lemmas: 
\begin{lemma}[Quantitative bijectivity] 
$\|G(\tilde{z}) - G(z)\| \geq \frac{1}{L^{l}_{\sigma}} \prod_{j=i}^L \sigma_{\min}(W_j) \|\tilde{z} - z\| $.  \\
\label{l:bijective}
\end{lemma} 
\begin{proof}
The proof proceeds by reverse induction on $l$. 
We will prove that 
$$\|h_i - \tilde{h}_{i}\| \geq \frac{1}{L^{l-i}_{\sigma}} \prod_{j=i}^L \sigma_{\min}(W_j) \|\tilde{z} - z\|$$ 
The claim trivial holds for $i=0$, so we proceed to the induction. 
Suppose the claim holds for $i$. Then, 
$$\|W_{i} h_i + b_i - (W_{i} \tilde{h}_{i} + b_i)\| \geq \frac{1}{\sigma_{\min}(W_i)} \|h_i - \tilde{h}_i\|$$ 
and 
$$ \|\sigma(W_{i} h_i + b_i) - \sigma(W_{i} \tilde{h}_{i} + b_i) \| \geq \frac{L_{\sigma}}{\sigma_{\min}(W_i)}  \|h_i - \tilde{h}_i\| $$   
by Lipschitzness of $\sigma^{-1}$. Since $h_{i-1} = \sigma(W_{i} h_i + b_i)$ and $\tilde{h}_{i-1} = \sigma(W_{i} h_i + b_i)$, 
$$\|h_{i-1} - \tilde{h}_{i-1}\| \geq \frac{1}{L^{l-(i-1)}_{\sigma}} \prod_{j=i-1}^L \sigma_{\min}(W_j) \|\tilde{z} - z\|$$ 
as we need. 
\end{proof} 

\begin{lemma} [Approximate inversion] Let $x \in \mathbb{R}^d$ be s.t. $\exists z, \|G_{\theta}(z) - x\| \leq \epsilon$. Then, there is a neural network $\NNN$ of size $O(ld^2)$, activation function $\sigma^{-1}$ and Lipschitz constant $L^l_{\sigma} \prod_{i=1}^l \frac{\sigma_{\max}(W_i)}{\sigma^2_{\min}(W_i)}$ which recovers a $\hat{z}$, s.t. 
$\|\hat{z} - z\| \leq \epsilon 2^{l} L^l_{\sigma} \prod_{i=1}^l \frac{1}{\sigma_{\min}(W_i)}  $  
\label{l:invert}
\end{lemma}  
\begin{proof} 
$\NNN$ will iteratively produce estimates $\hat{h}_i$, s.t. \\ 
(1) $\hat{h}_0 = x$ \\ 
(2) $\hat{h}_i = \sigma^{-1}(\mbox{argmin}_{h} \|W_i h + b_i - \sigma^{-1}(\hat{h}_{i-1})\|^2_2$) \\ 

We will prove by induction that $|h_i - \hat{h}_{i}| \leq \epsilon 2^i L^i_{\sigma} \prod_{j=1}^i \frac{1}{\sigma_{\min}(W_j)}$.   
The claim trivial holds for $i=0$, so we proceed to the induction. 
Suppose the claim holds for $i$. Then,  
\begin{align*} \min_{h} \|W_{i+1} h + b_{i+1} - \hat{h}_{i}\| &\leq \|W_{i+1} h_{i+1} + b_{i+1} - \sigma^{-1}(\hat{h}_i)\| \\
&=\|\sigma^{-1}(h_i) - \sigma^{-1}(\hat{h}_i)\| \\
&\leq L_{\sigma} \|h_i - \hat{h}_i\| \\   
&\leq \epsilon  2^i L^{i+1}_{\sigma} \prod_{j=1}^i \frac{1}{\sigma_{\min}(W_j)} 
\end{align*} 
where the last inequality holds by the inductive hypothesis, and the next-to-last one due to Lipschitzness of $\sigma^{-1}$. 

Hence, denoting $\tilde{h} = \mbox{argmin}_h  \|W_{i+1} h + b_{i+1} - \sigma^{-1}(\hat{h}_{i})\|^2_2 $, we have  
\begin{align*} \|W_{i+1} \tilde{h} - W{i+1} h_{i+1} )\| &= \|W_{i+1} \tilde{h} + b_{i+1} - \sigma^{-1}(\hat{h}_i) + \sigma^{-1}(\hat{h}_i) - W_{i+1} h_{i+1} - b_{i+1} \| \\
&\leq \|W_{i+1} \tilde{h} + b_{i+1} - \sigma^{-1}(\hat{h}_i)\| + \|\sigma^{-1}(\hat{h}_i) - W_{i+1} h_{i+1} - b_{i+1} \| \\ 
&= \|W_{i+1} \tilde{h} + b_{i+1} - \sigma^{-1}(\hat{h}_i)\| + \|\sigma^{-1}(\hat{h}_i) - \sigma^{-1}(h_i) \| \\ 
&\leq  \epsilon 2^{i+1} L^{i+1}_{\sigma} \prod_{j=1}^i \frac{1}{\sigma_{\min}(W_j)}  
\end{align*}  
This implies that 
$$\|W_{i+1} (\tilde{h} -  h_{i+1}) \| \leq  2 \epsilon  L^{i+1}_{\sigma} \prod_{j=1}^i \frac{1}{\sigma_{\min}(W_j)}$$ 
which in turns means 
$$ \|\tilde{h} -  h_{i+1}\| \leq   \epsilon 2^{i+1} L^{i+1}_{\sigma} \prod_{j=1}^{i+1} \frac{1}{\sigma_{\min}(W_j)} $$
which completes the claim. 

Turning to the size/Lipschitz constant of the neural network: all we need to notice is that 
$\hat{h}_i =\sigma^{-1}(W^{\dagger}_i (\hat{h}_{i-1} - b_{i}))$, which immediately implies the Lipschitzness/size bound by simple induction. 

\end{proof}

We also introduce a few tools to get a handle on functions that can be approximated efficiently by neural networks of small size/Lipschitz constant. 

\begin{lemma}[Composing Lipschitz functions] If $f:\mathbb{R}^{d_2} \to \mathbb{R}^{d_3}$ and $g:\mathbb{R}^{d_1} \to \mathbb{R}^{d_2}$ are $L, K$-Lipschitz functions respectively, then 
$f \circ g: \mathbb{R}^{d_1} \to \mathbb{R}^{d_3}$ is $L K $-Lipschitz. 
\label{l:compose}
\end{lemma} 
\begin{proof} 
The proof follows by definition essentially: 
$$ \|f(g(x)) - f(g(x'))\| \leq L \|g(x) - g(x')\| \leq L K \|x-x'\|$$
\end{proof}

\begin{lemma} [Calculating singular value decomposition approximately, \citep{demmel2007fast}] There is a neural network with size $O(n^3 \mbox{poly}(\log(1/\epsilon)))$ that given a symmetric matrix $A \in \mathbb{R}^{n \times n}$ with minimum eigenvalue gap $\min_{i \neq j} |\lambda_i - \lambda_j| \geq \delta$ and eigenvectors $\{u_i\}$ outputs $\{\tilde{u}_i, \tilde{\lambda}_i\}$ s.t. : \\ 
(1) $|\langle \tilde{u}_i, \tilde{u}_j \rangle | \leq \epsilon, \forall i \neq j$ and $\|\tilde{u}_i\| = 1 \pm \epsilon$. \\
(2) $|\tilde{u}_i - u_i| \leq \epsilon/\delta$, $|\tilde{\lambda}_i - \lambda_i| \leq \epsilon, \forall i$. \\
(Note the eigenvalue/eigenvector pairs for $A$ are unique since the minimum eigenvalue gap is non-zero). \\  
\label{l:eigendecomp}
\end{lemma} 
 
\begin{lemma} [Backpropagation, \citep{rumelhart1986learning}] Given a neural network $f: \mathbb{R}^m \to \mathbb{R}$ of depth $l$ and size $N$, there is a neural network of size $O(N+l)$ which calculates the gradient $\frac{\partial f}{\partial i}, i \in [m]$.   
\label{l:backprop}
\end{lemma} 

\subsection{Proof of Theorem \ref{t:mainadditive}}\label{sec:t:mainadditive}

We will proceed to prove the two parts one at a time. 

First, we prove the following lemma, which can be seen as a quantitative version of Laplace's method for evaluating integrals: 
\begin{lemma} [``Tail'' bound for integral at $z^*$]  
Let $x = G(z^*) + r$, for $\|r\| \leq 10 \beta \sqrt{d} \log d$, and $\|z^*\| \leq 10 \sigma \sqrt{d} \log d$.
The, for $\beta = O(\mbox{poly}(1/d))$, and 
$$\delta = 100 \beta \log(1/\beta)\frac{\sqrt{d}}{R}$$ 

it holds that
$$\int_{z:  \|z - z^*\| > \delta, z \in D_z} e^{f(z)} dz \leq \beta \int_{z \in D_z} e^{f(z)} dz$$
\label{l:ideal}
\end{lemma} 
\begin{proof}
Let's write 
\begin{equation} \int_{ z \in D_z} e^{f(z)} dz = \int_{z:  \|z - z^*\| \leq \delta} e^{f(z)} dz +  \int_{z:  \|z - z^*\| > \delta, z \in D_z} e^{f(z)} dz \label{eq:total} \end{equation}
To prove the claim of the Lemma, we will lower bound the first term, and upper bound the latter, from which the conclusion will follow. 

Consider the former term. Taylor expanding in a neighborhood around $z^*$ we have
$$f(z) = f(z^*) + (z-z^*)^{\top} \nabla^2 f(z^*) (z-z^*) \pm \frac{T}{\beta^2} \|z-z^*\|^3$$ 
where the first-order term vanishes since $z^*$ is a global optimum. Furthermore, $\nabla^2 f(z^*) \preceq 0$ for the same reason. Hence, by Taylor's theorem with remainder bounds, and using the fact that $e^x \geq 1+x$, we have
$$ \int_{z:  \|z - z^*\| \leq \delta} e^{f(z)} dz \geq \left(1 - \frac{T}{\beta^2} \delta^3\right)  f(z^*) \int_{z:  \|z - z^*\| \leq \delta} (z-z^*)^{\top} \nabla^2 f(z^*) (z-z^*) $$

The integral on the right is nothing more than the (unnormalized) cdf of a Gaussian with covariance matrix $(-\nabla^2 f(z^*))^{-1}$. 

Moreover, $-\nabla^2 f(z^*)$ is positive definite with smallest eigenvalue bounded by $\frac{R^2}{d \beta^2}$, since 
\begin{align*} -\nabla^2 f(z^*) &= \nabla^2 (\|z^*\|^2) + \frac{1}{\beta^2} \nabla^2 (\|G(z^*) - x\|^2) \\
&=  I + \frac{1}{\beta^2} \nabla^2 (\|G(z^*) - x\|^2)   \end{align*}
and 
$$\nabla^2 (\|G(z^*) - x\|^2) \succeq \sum_i \nabla G_i(z^*)\nabla^{\top}G_i(z^*) + \sum_i (G_i(z) - x_i) \nabla^2 G_i(z) $$ 
where $G_i$ is the $i$-th coordinate of $G$. We claim $ \sum_i \nabla G_i(z^*)\nabla^{\top}G_i(z^*) \succeq \frac{R^2}{d} I$, and $ \|\sum_i (G_i(z) - x_i) \nabla^2 G_i(z)\|_2 \lesssim \beta \sqrt{d} \log d$. The latter follows from the bound on $r$ and Cauchy-Schwartz. For the former, note that we have 
\begin{align*} v^{\top} \left(\sum_i \nabla G_i(z^*)\nabla^{\top}G_i(z^*)\right) v &= \sum_i \langle v,\nabla G_i(z^*) \rangle^2 \\ 
&= \sum_i \left(\lim_{\epsilon \to 0} \frac{G_i(z^* + \epsilon v) - G_i(z^*)}{\epsilon} \right)^2 \end{align*} 
By Lemma~\ref{l:bijective}, $\|G(z^* + \epsilon v) - G(z^*)\| \geq R \epsilon$, so $\exists i$, s.t. $|G_i(z^* + \epsilon v) - G_i(z^*)| \geq \frac{R \epsilon}{\sqrt{d}}$.
Hence, 
$$\sum_i \left(\lim_{\epsilon \to 0} \frac{G_i(z^* + \epsilon v) - G_i(z^*)}{\epsilon} \right)^2 \geq \frac{R^2}{d} $$
from which $\sum_i \nabla G_i(z^*)\nabla^{\top}G_i(z^*) \succeq \frac{R^2}{d}$ follows.

Using standard Gaussian tail bounds, since $\delta \geq \frac{\beta \log(1/\beta)}{R} \sqrt{d} $, we have
\begin{equation}  \int_{z:  \|z - z^*\| \leq \delta} (z-z^*)^{\top} \nabla^2 f(z^*) (z-z^*) \geq (1 - \beta) \mbox{det}(4 \pi (-\nabla^2 f(z^*)))^{1/2} \label{eq:bulk1} \end{equation}

We proceed to the latter term in~\cref{eq:total}. We have
\begin{align*} f(z^*) - f(z) &= \left| \frac{\|z^*\|^2 + \|G(z^*) - x\|^2}{\beta^2} - \|z\|^2 - \frac{\|G(z) - x\|^2}{\beta^2} \right| \\ 
&\geq \left|\frac{\|G(z^*) - x\|^2}{\beta^2} - \frac{\|G(z) - x\|^2}{\beta^2}\right| - \left|\|z\|^2 - \|z^*\|^2\right| \\ 
&=\left|\frac{\|G(z^*) - G(z^*) - r \|^2}{\beta^2} - \frac{\|G(z) - G(z^*) + r\|^2}{\beta^2}\right| - \left|\|z\|^2 - \|z^*\|^2\right|\\
&\stackrel{\mathclap{\circled{1}}}{\geq} \frac{(R \|z - z^*\| - \|r\|)^2}{\beta^2} - \frac{\|r\|^2}{\beta^2}  -  \left|\|z\|^2 - \|z^*\|^2\right| \\
&\stackrel{\mathclap{\circled{2}}}{\geq}  \frac{(R \|z - z^*\| - \|r\|)^2}{\beta^2} - \frac{\|r\|^2}{\beta^2}  -  \|z - z^*\|^2 - 2\|z - z^*\|\|z^*\|  \\
\end{align*}
where $\circled{1}$ follows from the definition of $x$, and $\circled{2}$ by triangle inequality. 

Note also that $\frac{1}{2} R \|z - z^*\| \geq \|r\| $
since $\delta \geq \frac{2 \|r\|}{R}$, which in turn implies

$$\frac{(R \|z - z^*\| - \|r\|)^2}{\beta^2} - \frac{\|r\|^2}{\beta^2}  -  \|z - z^*\|^2 - 2\|z - z^*\|\|z^*\|  \geq \frac{3}{16} \frac{R^2}{\beta^2}  \|z - z^*\| $$ 

Finally, $\beta^2 \leq \frac{R^2}{32} \|z^*\|$, so that it follows
$$ \|z - z^*\|^2 \left(\frac{3 R^2}{16 \beta^2} - 1\right) - 2\|z - z^*\|\|z^*\| \leq \frac{3 R^2}{32 \beta^2} \|z - z^*\|^2 $$ 
Putting these estimates together, we get $f(z) < f(z^*) - \frac{3 R^2}{32 \beta^2} \|z - z^*\|^2 $, which implies
$$\int_{z:  \|z - z^*\| > \delta} e^{f(z)} dz \leq  e^{f(z^*)}\int_{z:  \|z - z^*\| > \delta} e^{-\frac{3 R^2}{32 \beta^2} \|z - z^*\|^2} $$ 
The integral on the right is again the unnormalized cdf of a Gaussian with covariance matrix $\frac{3 R^2}{32 \beta^2} I$, so by Gaussian tail bounds again, and using that the smallest eigenvalue of $-\nabla^2 f(z^*)$
is lower-bounded by $\frac{R^2}{\beta^2}$ 
we have 
$$\int_{z:  \|z - z^*\| > \delta} e^{f(z)} \leq \beta e^{f(z^*)} \mbox{det} (4 \pi (-\nabla^2 f(z^*)))^{1/2} $$ 
as we wanted. 

Putting this together with~\cref{eq:bulk1}, we get the statement of the theorem. 

\end{proof} 

With this in mind, we can prove part (1) of our Theorem~\ref{thm:formal} restated below: 

\begin{theorem} 

There is a neural network $\NNN$ of size $\mbox{poly}(\frac{1}{\beta}, R, L_G, T, S, d)$ with Lipschitz constant $O(\mbox{poly}(R, L_G, d, T, S), 1/\beta^4)$ which given as input $x = G(z^*) + r$, for $\|r\| \leq 10 \beta \sqrt{d} \log d$, and $\|z^*\| \leq 10 \sqrt{d} \log d$ for $\beta = O(\mbox{poly}(1/d))$ outputs $\NNN(x)$, s.t. 
$$|\NNN(x) - \log p^{\beta}(x) | = O_{\mbox{poly}(d)}(\beta \log(1/\beta)) + \exp(-d)$$

\label{t:closetomanifold}
\end{theorem}

\begin{proof}

It suffices to approximate 
\begin{equation} \int_{z \in D_z} e^{f(z)} dz \label{eq:1} \end{equation}
up to a multiplicative factor of $1 \pm O_{\mbox{poly}(d)}(\beta \log(1/\beta))$, since the normalizing factor satisfies 
$$ \int_{x \in D_x} \int_{z \in D_z} e^{-\|z\|^2} e^{- \frac{\|G_{\theta}(z) - x\|^2}{\beta^2}}dz  = (1 \pm \exp(-d)) \mbox{det}(4 \pi I)^{-1/2} \mbox{det}(4 \pi/\beta^2 I)^{-1/2} $$

We will first present the algorithm, then prove that it: \\
(1) Approximates the integral as needed. \\   
(2) Can be implemented by a small, Lipschitz network as needed. \\ 

The algorithm is as follows: 
\begin{algorithm}[H]
\caption{Discriminator family with restricted approximability for degenerate manifold}
\begin{algorithmic}[1]
\STATE {\bf Parameters:} Matrices $E_1, E_2, \dots, E_r \in \mathbb{R}^{}$, matrices $W_1, W_2, \dots, W_l$.  
\STATE Let $\delta = \frac{100 \frac{L_G} \beta \log(1/\beta)}{R^2}$ and let $\mathcal{S}$ be the trivial $\beta^2$-net of the matrices with spectral norm bounded by $O(1/\beta^2)$. 
\STATE Let $\hat{z} = \NNN_{\mbox{inv}}(x)$ be the output of the ``invertor'' circuit of Lemma~\ref{l:invert}. 
\STATE Calculate $g = \nabla f(\hat{z}), H = \nabla^2 f(\hat{z})$ by the circuit implied in Lemma~\ref{l:backprop}. 
\STATE Let $M$ be the nearest matrix in $\mathcal{S}$ to $H$ and $E_i, i \in [r]$ be s.t. $M+E_i$ has $\Omega(\beta)$-separated eigenvalues. (If there are multiple $E_i$ that satisfy the separation condition, pick the smallest $i$.) 
\STATE Let $(e_i, \lambda_i)$ be approximate eigenvector/eigenvalue pairs of $H + E_i$ calculated by the circuit implied in Lemma~\ref{l:eigendecomp}.  
\STATE Approximate $I_i = \log \left(\int_{|c_i| \leq \delta} e^{c_i \langle e_i, g \rangle + \sum_i c^2_i \lambda_i} d c_i\right), i \in [r]$ by subdividing the interval $(0,\delta)$ into intervals of size $\beta^2$ and evaluating the resulting Riemannian sum instead of the integral. 
\STATE Output $\sum_i I_i$. 
\end{algorithmic}
\label{a:algdeg}
\end{algorithm}

First, we will show (1), namely that the Algorithm~\ref{a:algdeg} approximates the integral of interest. 
We'll use an approximate version of Lemma~\ref{l:ideal} -- with a slightly different division of where the ``bulk'' of the integral is located. 
As in Algorithm~\ref{a:algdeg}, let $\hat{z} = \NNN_{\mbox{inv}}(x)$ be the output of the ``invertor'' circuit of Lemma~\ref{l:invert}. 
and let 
$\delta = \frac{100 d \frac{L_G} \beta \log(1/\beta)}{R^2}$ and denote by $B$ the set $B = \{z: |\langle z - \hat{z}, e_i \rangle| \leq \delta \}$. 
Furthermore, let's define how the matrices $E_i, i \in [r]$ are to be chosen. Let $\mathcal{S}$ be an $\beta^2$-net of the matrices with spectral norm bounded by $O(1/\beta^2)$. We claim that there exist matrices $E_1, E_2, \dots, E_r$, $r = \Omega(d \log (1/\beta))$, s.t. if $M \in \mathcal{S}$, at least one of the matrices $M+E_i, i \in [r]$ has eigenvalues that are $\Omega(\beta)$-separated and $\|E_i\|_2 \leq \frac{\sqrt{d}}{\beta}$. Indeed, let  
let $E$ be a random Gaussian matrix with entrywise variance $\frac{1}{\beta^2}$. By Theorem 2.6 in \citep{nguyen2017random}, for any fixed matrix $A$, with probability $3/4$, $\min_i |\lambda_i(A+E) - \lambda_{i+1}(A+E)| = \Omega(\beta)$. The number of matrices in $\mathcal{S}$ is bounded by $2^{O(d \log(1/\beta))}$, so $E_i, i \in [r]$ exist by the probabilistic method.

We can write the integral of interest as 

\begin{align*}  \int_{z \in B} e^{f(z)} dz +  \int_{z \in \bar{B}} e^{f(z)} dz  \end{align*}
Note that 
\begin{align*} \|z - z^*\| &= \|z - \hat{z} + \hat{z} - z^*\| \\
&\geq \|z - \hat{z}\| - \|\hat{z} - z^*\| \\
&\geq \frac{4L_G}{R} \| \hat{z} - z\|\end{align*}  
This means that $\{z: \|z - z^*\| \leq \frac{4 L_G}{R} \delta \} \subseteq B$, so  
by Lemma~\ref{l:ideal}, we have 
$$\int_{z \in \bar{B}} e^{f(z)} \leq O(\beta) \int_{z} e^{f(z)} dz $$
which means that to prove the statement of the Theorem, it suffices for us to approximate $\int_{z \in B} e^{f(z)} dz$.  

Consider the former term first. By Taylor's theorem with remainder, expanding $f$ in a $\delta$-neighborhood near $\hat{z}$, we get 
$$f(z) = f(\hat{z}) + (z - \hat{z})^{\top} \nabla f(\hat{z}) + (z-\hat{z})^{\top} \nabla^2 f(\hat{z}) (z-\hat{z}) \pm \frac{T}{\beta^2} \|z-\hat{z}\|^3$$ 
For notational convenience, same as in Algorithm~\ref{a:algdeg}, let's denote by $H:= (z-\hat{z})^{\top} \nabla^2 f(\hat{z}) (z-\hat{z})$, and $g := \nabla f(\hat{z})$. Steps 5-8 effectively perform a change of basis in the eigenbasis of $H$ and evaluate the integral in this basis -- however to ensure Lipschitzness (which we prove later on), we will need to perturb $H$ slightly.

Let $M$ be the closest matrix to $\nabla^2 f(\hat{z})$ in the $\beta^2$-net $\mathcal{S}$ and let $e_i$ be approximate eigenvectors of $\tilde{H} = M + E_i$ in the sense of Lemma~\ref{l:eigendecomp}, s.t. the eigenvalues of $M+E_i$ are $\Omega(\beta)$-separated. 

Since $\|E_i\| \leq \frac{\sqrt{d}}{\beta}$, we have 
$$|(z-\hat{z})^{\top} \tilde{H} (z-\hat{z}) - (z-\hat{z})^{\top} H (z-\hat{z})| = O(\beta) $$
Hence, 
$$f(z) = f(\hat{z}) + (z - \hat{z})^{\top} \nabla f(\hat{z}) + (z-\hat{z})^{\top} \tilde{H} (z-\hat{z}) \pm \frac{T}{\beta^2} \|z-\hat{z}\|^3 \pm O(\beta)$$ 

Towards rewriting $f$ in the approximate basis $e_i$, let $z-\hat{z} = \sum_i c_i e_i$ for some scalars $c_i$. We have 
\begin{align*} f(z) &= f(\hat{z}) + \sum_i c_i \langle e_i, g \rangle + \sum_i c^2_i \lambda_i \pm \frac{T}{\beta^2} (\sum_i c^2_i)^{3/2} \end{align*}
By Taylor's theorem with remainder, $e^{x} = 1 + x \pm e \frac{x^2}{2}$, if $x < 1$.  
Hence, 
\begin{align*} \int_{z \in B} e^{f(z)} dz &= \left(1 \pm e d^{3/2} \frac{T}{\beta^2} \delta^3\right) e^{f(\hat{z})} 
\prod_i \int_{|c_i| \leq \delta} e^{c_i \langle e_i, g \rangle + \sum_i c^2_i \lambda_i} \end{align*}

Calculating the integral by subdividing $(0,\delta)$ into intervals of size $\beta^2$, and approximating the integral by the accompanying Riemannian 
sum,  and taking into account $|c_i| = O(\beta)$ and $\lambda_i, \|g\| = O(\frac{1}{\beta^2})$, we get a multiplicative approximation of 
$$\int_{|c_i| \leq \delta} e^{c_i \langle e_i, g \rangle + \sum_i c^2_i \lambda_i}$$
of the order $e^{\beta} = 1 + O(\beta)$, which is what we want. \\

We turn to implementing the algorithm by a small neural network with good Lipschitz constant. Both the Lipschitz constant and
the size will be handled by the composition Lemma \ref{l:compose} and analyzing each step of Algorithm~\ref{a:algdeg}. 
Steps 3 and 4 are handled by our helper lemmas: the invertor circuit by Lemma~\ref{l:invert} has Lipschitz constant $L_G$;  
calculating the Hessian $\nabla^2 f(\hat{z})$ can be performed by a polynomially sized neural network by Lemma~\ref{l:backprop} and since the third partial derivatives are bounded, so the output of this network is $O(\frac{T}{\beta^2})$-Lipschitz as well. We turn to the remaining steps. 

Proceeding to the eigendecomposition, by Lemma~\ref{l:eigendecomp}, we can perform an approximate eigendecomposition of $\tilde{H}$ with a network of size $O(\mbox{poly}(d))$ -- so we only need to handle the Lipschitzness. We will show that the result of Steps 5-6, the vectors $u_j$ and scalars $\lambda_j$ are Lipschitz functions of $H$.  

Suppose that $H$ and $H'$ are s.t. $\|H - H'\| \leq \beta^2$ first. Then, $H, H'$ are mapped to the same matrix $M$ in $\mathcal{S}$, $\{u_j\}$ and $\{u'_j\}$ are the eigenvectors of $H+E_i$ and $H'+E_i$ for some $i \in [r]$. First, by Weyl's theorem, since $H+E_i = M+ (H-M) + E_i$, the eigenvalues of $H+E_i$ are $\Omega(\beta) - \|H-M\| = \Omega(\beta)$-separated. Furthermore, since $H' + E_i = H + E_i + (H'-H)$, by Wedin's theorem, $\|u_j(H' + E_i) - u_j(H+E_i)\| = O\left( \|H-H'\|_2 \frac{1}{\beta}\right)$. 
If, on the other hand, $\|H - H'\| > \beta^2$, and $E_a, E_b$ are the perturbation matrices used for $H$ and $H'$,  
since the eigenvalues of $H + E_a$ are $ \Omega(\beta) $-separated, by Wedin's theorem,  
$$\|u_j(H' + E_b) - u_j(H+E_a)\| \leq O\left(\frac{1}{\beta}\left(\|H-H'\|_2 + \|E_b - E_a\|_2\right) \right) = O\left(\frac{1}{\beta} \|H-H'\|_2\right) $$  
so we get that the map to the vectors $u_j$ is $O(1/\beta)$-Lipschitz. A similar analysis shows that the map to the eigenvalues $\lambda_j$ is also Lipschitz. 

Finally, we move to calculating the integral in Step 7: the trivial implementation of the integral by a neural network has size $O(\mbox{poly}(\beta))$ 
and as a function of $e_i, g, \lambda_i$ is $O(1/\beta)$-Lipschitz, which proves the statement of the theorem.   

\end{proof}
Moving to Part (2) of Theorem~\ref{thm:formal}, we prove: 
\begin{theorem} 
Let $\NNN$ be the neural network $\NNN$ used in Theorem~\ref{t:closetomanifold}. The network additionally satisfies 
$\NNN(x) \leq \log p(x) + O_{R, L_G, d}(\beta \log(1/\beta)) + \exp(-d)$, $\forall x \in D$. 
\label{t:farfrommanifold}
\end{theorem} 
\begin{proof} 

Recalling the proof of Theorem~\ref{t:closetomanifold}, and reusing the notation there, 
we can express 
$$q(x) = \int_{z} \exp(f(z)) = \int_{z \in B} e^{f(z)} dz +  \int_{z \in \bar{B}} e^{f(z)} dz $$

Since the neural network $\NNN$ ignores the latter term, and we need only produce an upper bound, it suffices to show that $\NNN$ approximates  
$$ \int_{z:  \|z - \hat{z}\| \leq \delta} e^{f(z)} dz $$ 
up to a multiplicative factor of $1 - O(\beta \log(1/\beta))$. However, if we consider the proof of Theorem~\ref{t:closetomanifold}, we notice that the approximation consider there indeed serves our purpose: 
Taylor-expanding same as there, we have 
\begin{align*} \int_{z \in B} e^{f(z)} dz &= \left(1 \pm e d^{3/2} T_{\beta} \delta^3\right) e^{f(\hat{z})} 
\prod_i \int_{|c_i| \leq \delta} e^{c_i \langle e_i, g \rangle + \sum_i c^2_i \lambda_i} \end{align*}
This integral can be evaluated in the same manner as in Theorem~\ref{t:closetomanifold}, as our bound on 
$T_{\beta}$ holds universally on neighborhood of radius $D_x$. 
 
\end{proof} 

Finally, part (3) follows easily from (1) and (2): 
\begin{proof}[Proof of Part 3 of Theorem~\ref{thm:formal}] 
For points $x$, s.t. $\nexists z, \|G(z) - x\| \leq 10 \beta \sqrt{d} \log d$, it holds that $p(x) = O(\exp(-d))$. On the other hand, by the Lipschitzness of $\NNN$, we have $\|\NNN(x)\| = O_{\frac{1}{\beta^4}}(\|x\|)$. Since $x \in D_x$ implies $\|x\| = O(\mbox{poly}(d))$ the claim follows.      
\end{proof}

%% file: cleantex-new/appendix-experiment-iclr.tex
\section{Experiments on Synthetic 2d Datasets: Unit Circle}
\label{appendix:toy-experiment}

\begin{figure}[h!]
  \centering
  \begin{subfigure}[b]{0.29\textwidth}
    \includegraphics[width=\textwidth]{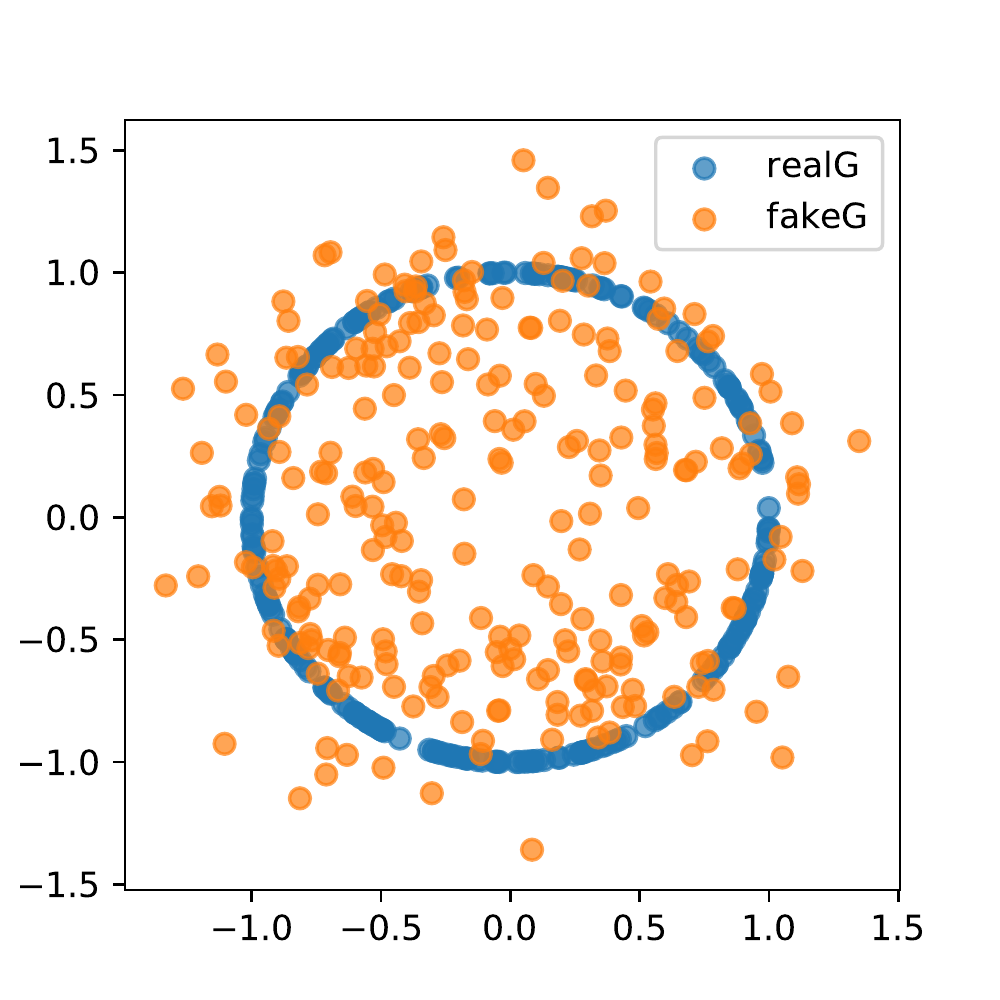}
    \caption{\small Iteration 500.}
    \label{figure:circle1}
  \end{subfigure}
  ~
  \begin{subfigure}[b]{0.29\textwidth}
    \includegraphics[width=\textwidth]{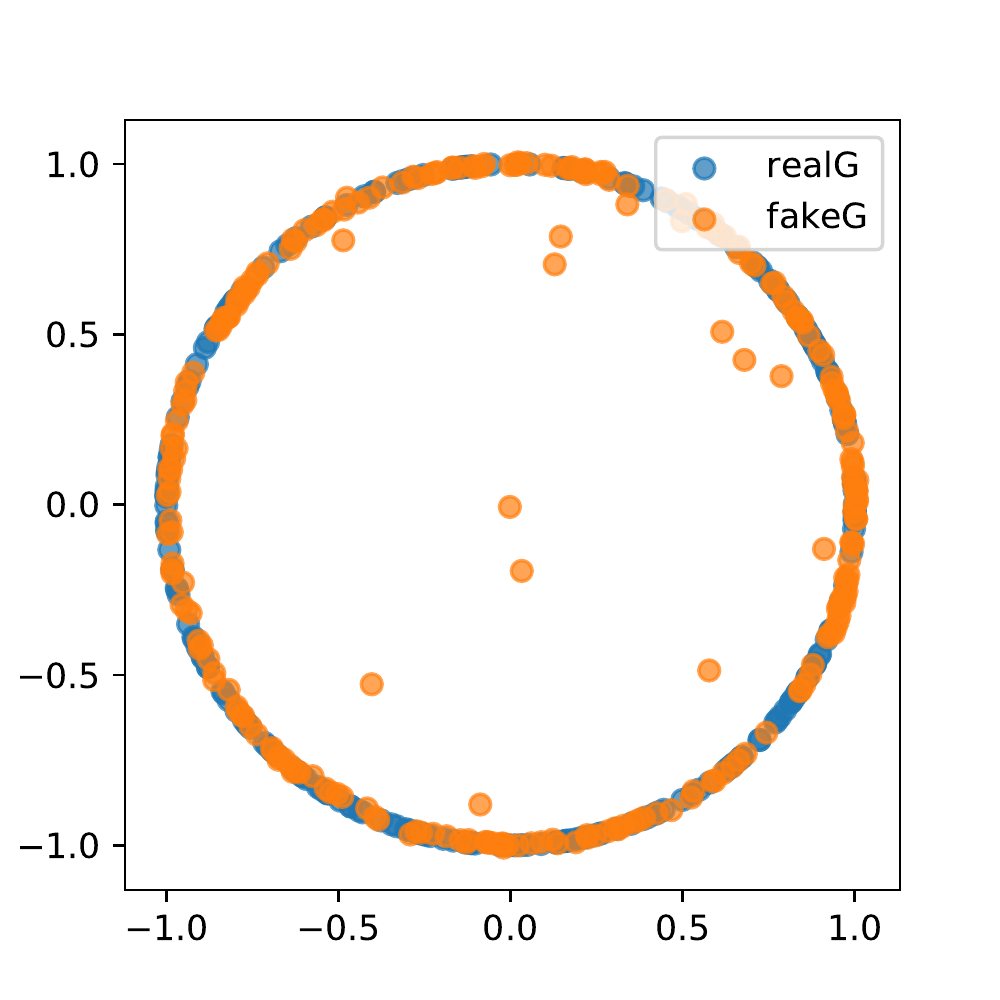}
    \caption{\small Iteration 10000.}
    \label{figure:circle2}
  \end{subfigure}
  ~
  \begin{subfigure}[b]{0.35\textwidth}
    \includegraphics[width=\textwidth]{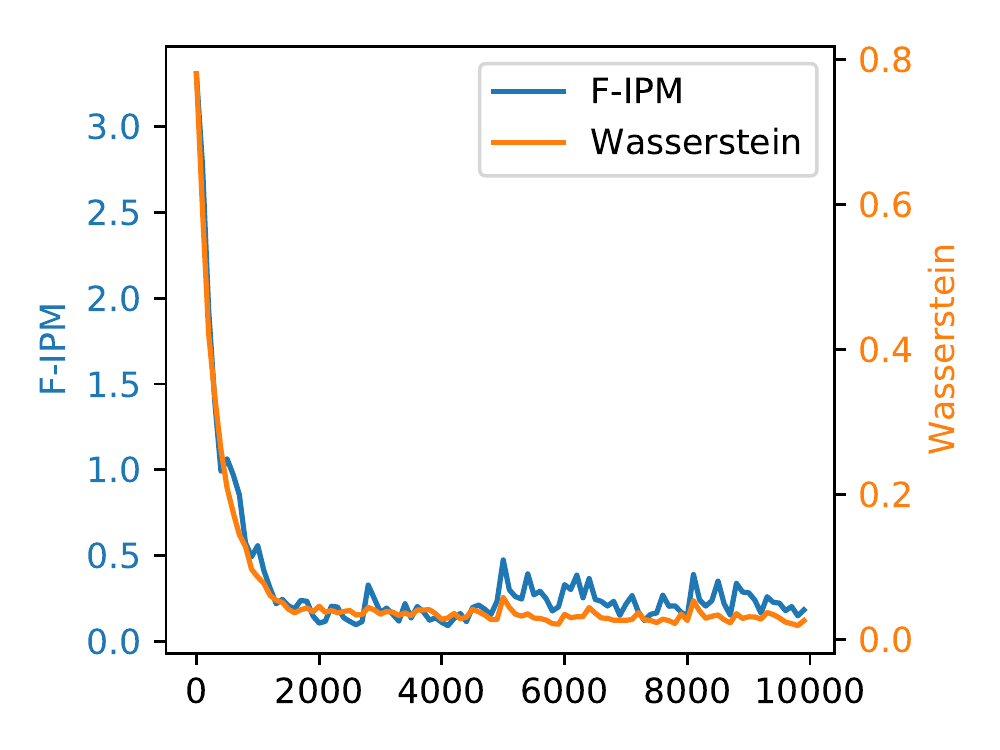}
    \caption{\small Comparing IPM and Wasserstein.}
    \label{figure:circle3}
  \end{subfigure}
  \caption{\small Experiments on the unit circle dataset. The neural
    net IPM, the Wasserstein distance, and the sample quality are
    correlated along training. (a)(b): Sample batches from the ground
    truth and the learned generator at iteration 500 and 10000. (c):
    Comparing the F-IPM and the Wasserstein distance. RealG and fakeG
    denote the ground truth generator and the learned generator,
    respectively.}
  \label{figure:circle}
\end{figure}



\section{Experiments on Invertible Neural Net Generators}
\label{sec:exp}
We further perform synthetic WGAN experiments with invertible neural
net generators (cf. Section~\ref{section:invertible-generator}) and
discriminators designed with restricted approximability
(Lemma~\ref{lemma:logp-neural-network}). In this case, the
invertibility guarantees that the KL divergence can be computed, and
our goal is to demonstrate that the empirical IPM $W_{\cF}(p,q)$ is
well correlated with the KL-divergence between $p$ and $q$ on
synthetic data for various pairs of $p$ and $q$ (The true
distribution $p$ is generated randomly from a ground-truth neural net,
and the distribution $q$ is learned using various algorithms or
perturbed version of $p$.)


\subsection{Setup}
\paragraph{Data}
The data is generated from a ground-truth invertible neural net
generator (cf. Section~\ref{section:invertible-generator}),
i.e. $X=G_\theta(Z)$, where $G_\theta:\R^d\to\R^d$ is a $\ell$-layer
layer-wise invertible feedforward net, and $Z$ is a spherical
Gaussian. We use the Leaky ReLU with negative slope 0.5 as the
activation function $\sigma$, whose derivative and inverse can be very
efficiently computed. The weight matrices of the layers are set to be
well-conditioned with singular values in between $0.5$ to $2$.

\sloppy We choose the discriminator architecture according to the
design with restricted approximability guarantee
(Lemma~\ref{lemma:logp-neural-network},~\cref{eqn:F-1}~\cref{eqn:F-2}). As
$\log\sigma^{-1'}$ is a piecewise constant function that is not
differentiable, we instead model it as a trainable one-hidden-layer
neural network that maps reals to reals. We add constraints on all the
parameters in accordance with
Assumption~\ref{assumption:invertible-generator}.

\paragraph{Training}
To train the generator and discriminator networks, we generate
stochastic batches (with batch size 64) from both the ground-truth
generator and the trained generator, and solve the min-max problem in
the Wasserstein GAN formulation. We perform 10 updates of the
discriminator in between each generator step, with various
regularization methods for discriminator training (specified
later). We use the RMSProp optimizer~\citep{TielemanHi12} as our update
rule.

\paragraph{Evaluation metric}
We evaluate the following metrics between the true and learned
generator.
\begin{enumerate}[(1)]
\item The KL divergence. As the density of our invertible neural net
  generator can be analytically computed, we can compute their KL
  divergence from empirical averages of the difference of the log densities: 
  \begin{equation*}
    \what{\dkl}(p^\star,p) = \E_{X\sim\what{p^\star}^n}[\log
    p^\star(X)-\log p(X)],
  \end{equation*}
  where $p^\star$ and $p$ are the densities of the true generator and
  the learned generator. We regard the KL divergence as the
  ``correct'' and rather strong criterion for distributional
  closeness.
\item The training loss (IPM $W_{\cF}$ train). This is the
  (unregularized) GAN loss during training. Note: as typically in the
  training of GANs, we balance carefully the number of steps for
  discriminator and generators, the training IPM is potentially very
  far away from the true $W_{\cF}$ (which requires sufficient training of
  the discriminators).
\item The neural net IPM ($W_{\cF}$ eval). We report once in a while a
  separately optimized WGAN loss in which the learned generator is
  held fixed and the discriminator is trained from scratch to
  optimality. Unlike the training loss, here the discriminator is
  trained in norm balls but with no other regularization. By doing
  this, we are finding $f\in\cF$ that maximizes the contrast and we
  regard the $f$ found by stochastic optimization an approximate
  maximizer, and the loss obtained an approximation of $W_{\cF}$.
\end{enumerate}

Our theory shows that for our choice of $\mc{G}$ and $\mc{F}$, WGAN is
able to learn the true generator in KL divergence, and the $\cF$-IPM
(in evaluation instead of training) should be indicative of the KL
divergence. We test this hypothesis in the following experiments.

\vspace{0.1cm}

\subsection{Convergence of generators in KL divergence}
\label{section:experiment-convergence}
\begin{figure}[h!]
  \centering
  \includegraphics[width=0.9\columnwidth]{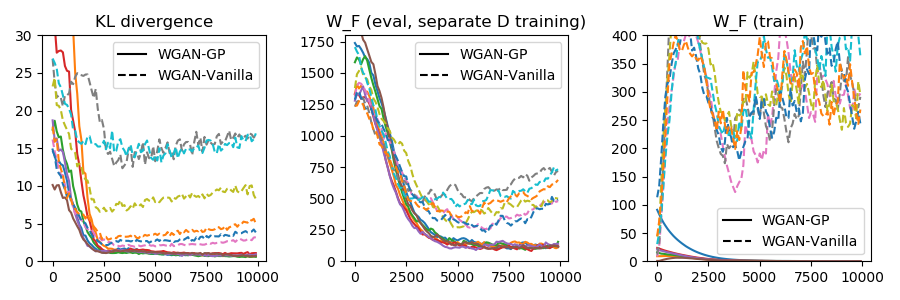}
  \caption{\small Learning an invertible neural net generator on
    synthetic data.  The x-axis in all the graphs indicates the number
    of steps. The left-most figure shows the KL-divergence between the
    true distribution $p$ and learned distribution $q$ at different
    steps of training, the middle the estimated IPM (evaluation)
    between $p$ and $q$, and the right one the training loss. We see
    that the estimated IPM in evaluation correlates well with the
    KL-divergence. Moving average is applied to all curves. }
  \label{figure:conjoined}
\end{figure}

In our first experiment, $G$ is a two-layer net in $d=10$
dimensions. Though the generator is only a shallow neural net, the
presence of the nonlinearity makes the estimation problem
non-trivial. We train a discriminator with the architecture specified
in Lemma~\ref{lemma:logp-neural-network}), using either Vanilla WGAN
(clamping the weight into norm balls) or
WGAN-GP~\citep{GulrajaniAhArDuCo17} (adding a gradient penalty). We fix
the same ground-truth generator and run each method from 6 different
random initializations. Results are plotted in
Figure~\ref{figure:conjoined}.

Our main findings are two-fold:
\begin{enumerate}[(1)]
\item WGAN training with discriminator design of restricted
  approximability is able to learn the true distribution in KL
  divergence. Indeed, the KL divergence starts at around 10 - 30 and
  the best run gets to KL lower than 1. As KL is a rather strong
  metric between distributions, this is strong evidence that GANs are
  finding the true distribution and mode collapse is not happening.
\item The $W_{\cF}$ (eval) and the KL divergence are highly correlated
  with each other, both along each training run and across different
  runs. In particular, adding gradient penalty improves the
  optimization significantly (which we see in the KL curve), and this
  improvement is also reflected by the $W_{\cF}$ curve. Therefore the
  quantity $W_{\cF}$ can serve as a good metric for monitoring
  convergence and is at least much better than the training loss
  curve.
\end{enumerate}

To test the necessity of the specific form of the discriminator we
designed, we re-do the same experiment with vanilla fully-connected
discriminator nets. Results (in
Appendix~\ref{appendix:experiment-vanilla}) show that IPM with vanilla
discriminators also correlate well with the KL-divergence. This is not
surprising from a theoretical point of view because a standard
fully-connected discriminator net (with some over-parameterization) is
likely to be able to approximate the log density of the generator
distributions (which is essentially the only requirement of
Lemma~\ref{thm:logp}.)

For this synthetic case, we can see that the inferior performance in
KL of the WGAN-Vanilla algorithm doesn't come from the statistical
properties of GANs, but rather the inferior training performance in
terms of the convergence of the IPM. We conjecture similar phenomenon
occurs in training GANs with real-life data as well.

\subsection{Perturbed generators}
\begin{figure}[h!]
  \centering
  \includegraphics[width=0.4\columnwidth]{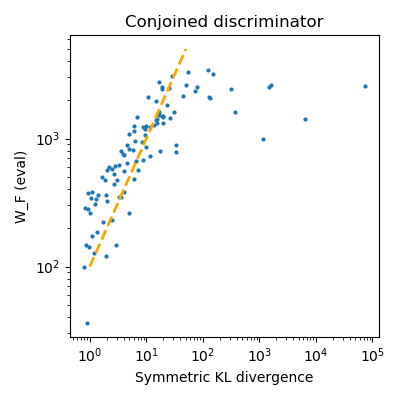}
  \caption{Scatter plot of KL divergence and neural net IPM on
    perturbed generator pairs. Correlation between
    $\log(\dkl(p\|q)+\dkl(q\|p))$ and 
    $\log\df$ is 0.7315. Dashed line is
    $\df(p,q)=100(\dkl(p\|q)+\dkl(q\|p))$.}
  \label{figure:correlation}
\end{figure}

In this section, we remove the effect of the optimization and directly
test the correlation between $p$ and its perturbations.  We compare
the KL divergence and neural net IPM on pairs of {\it perturbed
  generators}.  In each instance, we generate a pair of generators
$(G,G')$ (with the same architecture as above), where $G'$ is a
perturbation of $G$ by adding small Gaussian noise.  We compute the KL
divergence and the neural net IPM between $G$ and $G'$. To denoise the
unstable training process for computing the neural net IPM, we
optimize the discriminator from 5 random initializations and pick the
largest value as the output.

As is shown in Figure~\ref{figure:correlation}, there is a clear
positive correlation between the (symmetric) KL divergence and the
neural net IPM.  In particular, majority of the points fall around the
line $\df=100\dkl$, which is consistent with our theory that the
neural net distance scales linearly in the KL divergence.  Note that
there are a few outliers with large KL. This happens mostly
due to the perturbation being accidentally too large so that the
weight matrices become poorly conditioned -- in the context of our theory,
they fall out of the good constraint set as defined in
Assumption~\ref{assumption:invertible-generator}.

\subsection{Experiments with vanilla discriminator}
\label{appendix:experiment-vanilla}
\subsubsection{Convergence of generators in KL divergence}
\begin{figure}[h!]
  \centering
  \includegraphics[width=0.9\columnwidth]{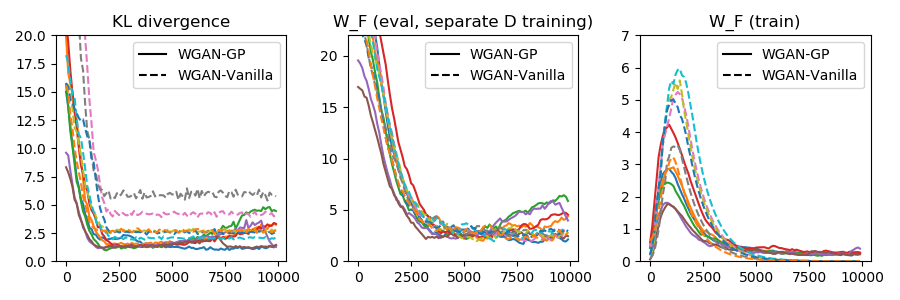}
  \caption{\small Learning an invertible neural net generator on
    synthetic data with vanilla fully-connected discriminator nets.
    The x-axis in all the graphs indicates the number
    of steps. The left-most figure shows the KL-divergence between the
    true distribution $p$ and learned distribution $q$ at different
    steps of training, the middle the estimated IPM (evaluation)
    between $p$ and $q$, and the right one the training loss. We see
    that the estimated IPM in evaluation correlates well with the
    KL-divergence. Moving average is applied to all curves.}
  \label{figure:vanilla}
\end{figure}

We re-do the experiments of
Section~\ref{section:experiment-convergence} with vanilla
fully-connected discriminator nets. We use a three-layer net with
hidden dimensions 50-10, which has more parameters than the
architecture with restricted approximability. Results are plotted in
Figure~\ref{figure:vanilla}. We find that the generators also converge
well in the KL divergence, but the correlation is slightly weaker
than the setting with restricted approximability (correlation
still presents along each training run but weaker across different
runs). This suggests that vanilla discriminator structures might be
practically quite satisfying for getting a good generator, though
specific designs may help improve the quality of the distance
$W_{\cF}$.

\subsubsection{Perturbed generators}
\begin{figure}[h!]
  \centering
  \includegraphics[width=0.4\columnwidth]{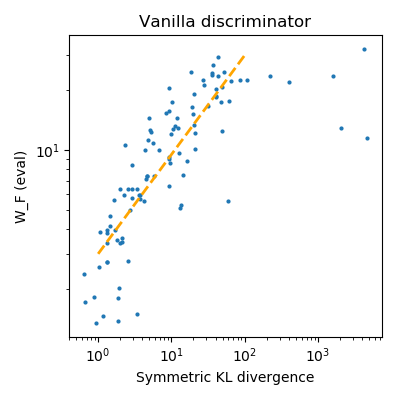}
  \caption{Scatter plot of KL divergence and neural net IPM (with
    vanilla discriminators) on
    perturbed generator pairs. Correlation between
    $\log(\dkl(p\|q)+\dkl(q\|p))$ and 
    $\log W_{\cF}$ is 0.7489. Dashed line is
    $\df(p,q)=3\sqrt{\dkl(p\|q)+\dkl(q\|p)}$.}
  \label{figure:correlation-vanilla}
\end{figure}

Correlation between KL and neural net IPM is computed with vanilla
fully-connected discriminators and plotted in
Figure~\ref{figure:correlation-vanilla}. The correlation (0.7489) is
roughly the same as for discriminators with restricted approximability
(0.7315).
